\documentclass[10pt,twocolumn,letterpaper]{article}

\usepackage{cvpr}
\usepackage{times}  %Required
\usepackage{helvet}  %Required
\usepackage{courier}  %Required
\usepackage{url}  %Required
\usepackage{graphicx}  %Required
\usepackage{times}
\usepackage{epsfig}
\usepackage{amsmath}
\usepackage{amssymb}
\usepackage[noend]{algpseudocode}
\usepackage{algorithm}
\usepackage{tabularx}
\usepackage{array}
\usepackage{booktabs}
\usepackage{subfigure}
\usepackage{multirow}
\usepackage{arydshln}
\usepackage{tablefootnote}
\usepackage[utf8]{inputenc}
\usepackage[english]{babel}
\usepackage{amsthm}
\usepackage{scrextend}
\newtheorem{theorem}{Theorem}

\newtheorem{proposition}[theorem]{Proposition}
\newtheorem{definition}[theorem]{Definition}
\def\mb{\mathbf}

\def\etal{\emph{et al.} }
\def\eg{\emph{e.g.}}
\def\ie{\emph{i.e.}}
\def\wrt{\emph{w.r.t.} }
\usepackage[breaklinks=true,bookmarks=false]{hyperref}

\cvprfinalcopy % *** Uncomment this line for the final submission

\def\cvprPaperID{601} % *** Enter the CVPR Paper ID here

\begin{document}

%%%%%%%%% TITLE
% \title{Pruning Networks Using Neuron Importance Score Propagation}
\title{NISP: Pruning Networks using Neuron Importance Score Propagation}

% \title{Supplementary Materials of NISP: Pruning Networks \\ using Neuron Importance Score Propagation}

% \and

\author{Ruichi Yu$^{1}$ ~~~Ang  Li$^3$\thanks{This work was done while the author was at University of Maryland.}
~~~ Chun-Fu Chen$^2$~~~ Jui-Hsin Lai$^5$\thanks{This work was done while the author was at IBM.}  ~~~ Vlad I. Morariu$^{4*}$ \\
~~~ Xintong Han$^1$~~~ Mingfei Gao$^1$ ~~~Ching-Yung Lin$^6\dag$ ~~~ Larry S. Davis$^1$\\
$^1$University of Maryland, College Park ~~~~~~~~~~$^2$IBM T. J. Watson Research  \\
$^3$DeepMind  ~~~~~~~~~~$^4$Adobe Research ~~~~~~~~~~$^5$JD.com~~~~~~~~~~$^6$Graphen.ai\\
{\tt\small {\{richyu, xintong, mgao, lsd\}@umiacs.umd.edu}, anglili@google.com
}
\\
{\tt\small { chenrich@us.ibm.com}, larry.lai@jd.com, morariu@adobe.com, cylin@graphen.ai
}
}

\maketitle
%\thispagestyle{empty}

%%%%%%% ABSTRACT
\begin{abstract}
To reduce the significant redundancy in deep Convolutional Neural Networks (CNNs), most existing methods prune neurons by only considering statistics of an individual layer or two consecutive layers (e.g., prune one layer to minimize the reconstruction error of the next layer), ignoring the effect of error propagation in deep networks.
In contrast, we argue that it is essential to prune neurons in the entire neuron network jointly based on a unified goal: minimizing the reconstruction error of important responses in the ``final response layer" (FRL), which is the second-to-last layer before classification, for a pruned network to retrain its predictive power.
Specifically, we apply feature ranking techniques to measure the importance of each neuron in the FRL, and formulate network pruning as a binary integer optimization problem and derive a closed-form solution to it for pruning neurons in earlier layers. Based on our theoretical analysis, we propose the Neuron Importance Score Propagation (NISP) algorithm to propagate the importance scores of final responses to every neuron in the network. The CNN is pruned by removing neurons with least importance, and then fine-tuned to retain its predictive power.
NISP is evaluated on several datasets with multiple CNN models and demonstrated to achieve significant acceleration and compression with negligible accuracy loss. 
\end{abstract}

\section{Introduction}

% Convolutional Neural Networks (CNNs) have achieved state-of-the-art performance on many computer vision tasks, such as image classification, object detection and autonomous driving.
% However, 
CNNs require a large number of parameters and high computational cost in both training and testing phases. 
Recent studies have investigated the significant redundancy in deep networks~\cite{PredictingParameters}
% Some methods reduce the computational cost of a CNN by approximating the redundant structures by less complex ones \cite{DentonLeCun,SeperableFilter,Nonlinear,PerforatedCNN,Tucker}, 
and reduced the number of neurons and filters \cite{random,DeepCompress,pruneweigth,thinet} by pruning the unimportant ones.
% The advantage of the latter is that they keep CNN architectures unchanged (only change hyper-parameters, \eg, the number of neurons), which is generic for different architectures. 
However, most current approaches that prune neurons and filters consider only the statistics of one layer (\eg, prune neurons with small magnitude of weights \cite{pruneweigth,DeepCompress}), or two consecutive layers \cite{thinet} to determine the ``importance" of a neuron. These methods prune the ``least important" neurons layer-by-layer either independently \cite{DeepCompress} or greedily \cite{pruneweigth, thinet}, without considering all neurons in different layers jointly.

One problem with such methods is that neurons deemed unimportant in an early layer can, in fact, contribute significantly to responses of important neurons in later layers. Our experiments (see Sec.\ref{LBL}) reveal that greedy layer-by-layer pruning leads to significant reconstruction error propagation, especially in deep networks, which indicates the need for a global measurement of neuron importance across different layers of a CNN.

\begin{figure}[t]
\begin{center}
   \includegraphics[width=\linewidth]{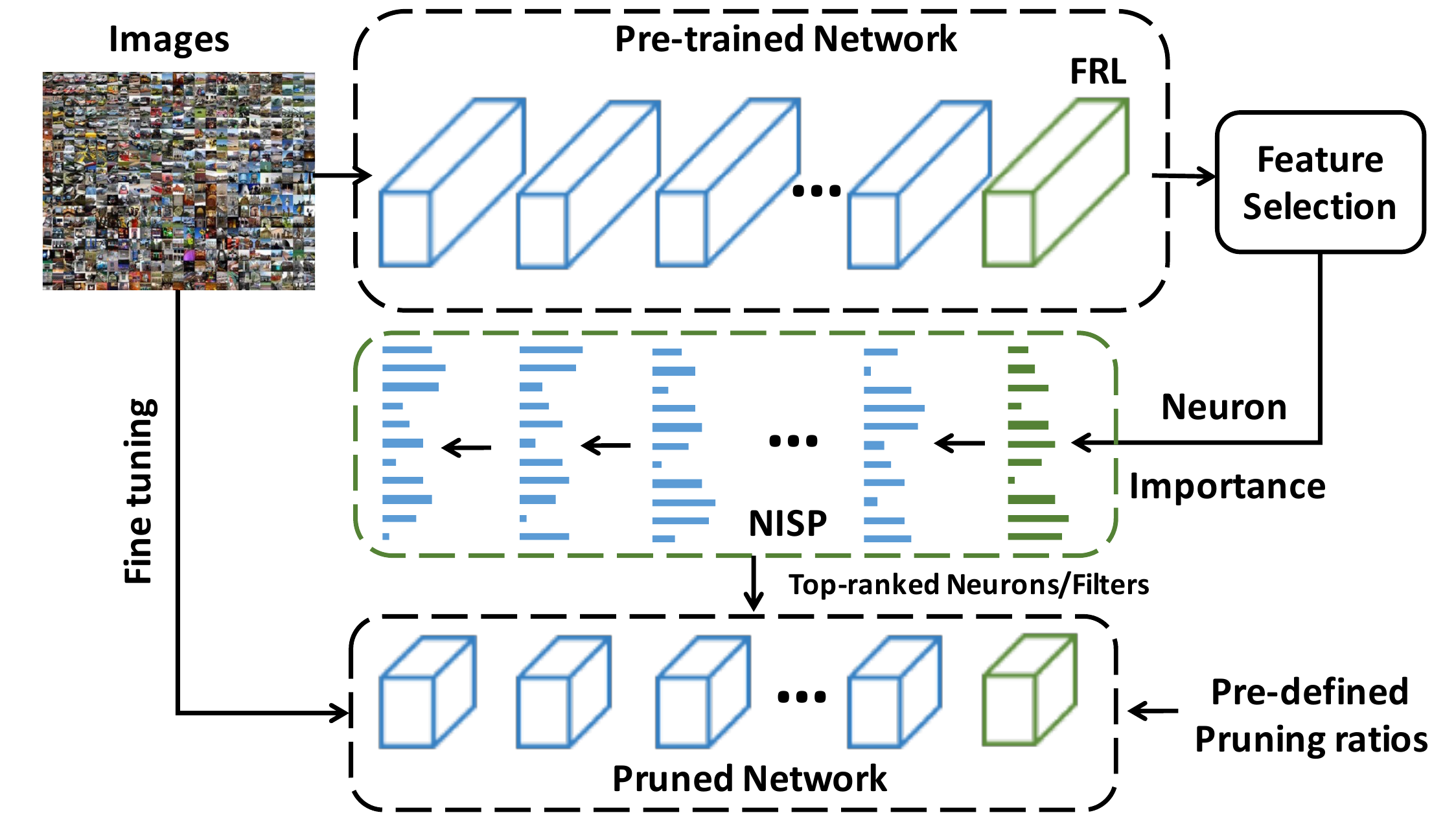}
\end{center}
   \caption{We measure the importance of neurons in the final response layer (FRL), and derive Neuron Importance Score Propagation (NISP) to propagate the importance to the entire network. Given a pre-defined pruning ratio per layer, we prune the neurons/filters with lower importance score. We finally fine-tune the pruned model to recover its predictive accuracy.}
\label{fig:spe}
\end{figure}

To address this problem, we argue that it is essential for a pruned model to retain the most important responses of the second-to-last layer before classification (``final response layer" (FRL)) to retrain its predictive power, since those responses are the direct inputs of the classification task (which is also suggested by feature selection methods, \eg, \cite{Roffo_2015_ICCV}). We define the importance of neurons in early layers based on a \textbf{unified goal}: \emph{minimizing the reconstruction errors of the responses produced in FRL.} 
We first measure the importance of responses in the FRL by treating them as features and applying some feature ranking techniques (\eg, \cite{Roffo_2015_ICCV}), then propagate the importance of neurons backwards from the FRL to earlier layers. We prune only nodes which have low propagated importance (\ie, those whose removal does not result in large propagated error).  From a theoretical perspective, we formulate the network pruning problem as a binary integer programming objective that minimizes the weighted $\ell^1$ distance (proportional to the importance scores) between the original final response and the one produced by a pruned network. 
% The distance weights are proportional to the importance scores of final responses given by a feature ranking algorithm. 
We obtain a closed-form solution to a relaxed version of this objective to infer the importance score of every neuron in the network. 
Based on this solution, we derive the \textit{Neuron Importance Score Propagation} (NISP) algorithm, which computes all importance scores recursively, using only one feature ranking of the final response layer and one backward pass through the network as illustrated in Fig.~\ref{fig:spe}.

The network is then pruned based on the inferred neuron
importance scores and fine-tuned to retain its predictive capability.
We treat the pruning ratio per layer as a pre-defined hyper-parameter, which can be determined based on different needs of specific applications (\eg, FLOPs, memory and accuracy constraints). The pruning algorithm is generic, since feature ranking can be applied to any layer of interest and the importance scores can still be propagated. In addition, NISP is not hardware specific. Given a pretrained model, NISP outputs a smaller network of the same type, which can be deployed on the hardware devices designed for the original model.
% Meanwhile, since we directly reduce the input volume of each layer, the computations of all layers include pooling and activation layers are reduced so that we can achieve a full-network acceleration and compression. 

We evaluate our approach on MNIST \cite{lenet}, CIFAR10 \cite{CIFAR10} and ImageNet \cite{imagenet_cvpr09} using multiple standard CNN architectures such as LeNet \cite{lenet}, AlexNet \cite{Alexnet}, GoogLeNet \cite{googlenet} and ResNet \cite{resnet}. 
Our experiments show that CNNs pruned by our approach outperform those with the same structures but which are either trained from scratch or randomly pruned. We demonstrate that our approach outperforms magnitude-based and layer-by-layer pruning. 
%We conduct further experiments on AlexNet and GoogLeNet to analyze the redundancy of different layers and provide guidance for pruning a CNN.  
% Our method is orthogonal to most of the existing methods. Since the output of our approach has the same general structure as standard CNNs, but with fewer neurons and better weights, existing approaches can build on our output. Nevertheless, w
A comparison of the theoretical reduction of FLOPs and number of parameters of different methods shows that our method achieves faster full-network acceleration and compression with lower accuracy loss, \eg, our approach loses 1.43\% accuracy on Alexnet and reduces FLOPs by 67.85\% while Figurnov \etal \cite{PerforatedCNN} loses more (2\%) and reduces FLOPs less (50\%).  With almost zero accuracy loss on ResNet-56, we achieve a 43.61\% FLOP reduction, significantly higher than the 27.60\% reduction by Li \etal \cite{pruneweigth}.

%by keeping smaller accuracy loss (1.45\% ours vs. 2.00\%\cite{PerforatedCNN}) on AlexNet, our method reduces 67.85\% FLOPs while \cite{PerforatedCNN} reduces only 50\%;

\subsection{Contribution} We introduce a generic network pruning algorithm which formulates the pruning problem as a binary integer optimization and provide a closed-form solution based on final response importance. We present NISP to efficiently propagate the importance scores from final responses to all other neurons. Experiments demonstrate that NISP leads to full-network acceleration and compression for all types of layers in a CNN with small accuracy loss.

\section{Related Work}
% Despite their impressive predictive power on a wide range of tasks \cite{xu1,xu2,xu3,yu1,yu2,yu3,yu4,yu5,yu6,yu7}, the redundancy in the parameterization of deep learning models has been studied and demonstrated \cite{PredictingParameters}.
There has been recent interest in reducing the redundancy of deep CNNs to achieve acceleration and compression. 
In \cite{PredictingParameters} the redundancy in the parameterization of deep learning models has been studied and demonstrated.
Cheng \emph{et al.} \cite{Circulant} exploited properties of structured matrices and used circulant matrices to represent FC layers, reducing storage cost. 
Han \emph{et al.} \cite{DeepCompress} studied the weight sparsity and compressed CNNs by combining pruning, quantization, and Huffman coding. Sparsity regularization terms have been use to learn sparse CNN structure in \cite{lasso,SSL,learning}. Miao \emph{et al.} \cite{miao-icde} studied network compression based on float data quantization for the purpose of massive model storage.

To accelerate inference in convolution layers, Jaderberg \emph{et al.} \cite{SeperableFilter} constructed a low rank basis of filters that are rank-1 in the spatial domain by exploiting cross-channel or filter redundancy.
Liu \emph{et al.} \cite{slimLiu} imposed a scaling factor in the training process and facilitated one channel-level pruning.
Figurnov \emph{et al.} \cite{PerforatedCNN} speeded up the convolutional layers by skipping operations in some spatial positions, which is based on loop perforation from source code optimization.
In \cite{DentonLeCun,Nonlinear,Tucker}, low-rank approximation methods have been utilized to speed up convolutional layers by decomposing the weight matrix into low-rank matrices. 
Molchanov \emph{et al.} \cite{Nvidia} prune CNNs based on Taylor expansion.

Focusing on compressing the fully connected (FC) layers, Srinivas \emph{et al.} \cite{Datafree} pruned neurons that are similar to each other.
Yang \emph{et al.} \cite{fry} applied the ``Fastfood" transform to reparameterize the matrix-vector multiplication of FC layers. 
Ciresan \emph{et al.} \cite{random} reduced the parameters by randomly pruning neurons.
Chen \emph{et al.} \cite{hash} used a low-cost hash function to randomly group connection weights into hash buckets and then fine-tuned the network with back-propagation.
Other studies focused on fixed point computation rather than exploiting the CNN redundancy \cite{precision,binary}. Another work studied the fundamental idea about knowledge distillation \cite{Distilling}. Wu \textit{et al.} \cite{Zuxuan} proposed to skip layers for speeding up inference. Besides the above work which focuses on network compression, other methods speedup deep network inference by refining the pipelines of certain tasks \cite{faster,yu1,yu3, ssd}.
Our method prunes a pre-trained network and requires a fast-converging fine-tuning process, rather than re-training a network from scratch.
To measure the importance of neurons in a CNN, the exact solution is very hard to obtain given the complexity of nonlinearity. Some previous works \cite{brain2,brain3,brain} approximate it using 2nd-order Taylor expansion. Our work is a different approximation based on the Lipschitz continuity of a neural network.

Most similar to our approach, Li \emph{et al.} \cite{pruneweigth} pruned filters by their weight magnitude. Luo \emph{et al.} \cite{thinet} utilized statistics information computed from the next layer to guide a greedy layer-by-layer pruning. In contrast, we measure neuron importance based not only on a neuron's individual weight but also the properties of the input data and other neurons in the network. Meanwhile, instead of pruning layer-by-layer in greedy fashion under the assumption that one layer can only affect its next layer, which may cause error propagation, we measure the importance across the entire network by propagating the importance from the final response layer.

\section{Our Approach}
% We follow a straightforward strategy to reduce Convolutional Neural Networks: pruning kernels of convolution layers and neurons in fully connected layers. Trivial methods like random pruning or changing the number of neurons and kernels arbitrarily may result in huge degradation of the predictive power. 
% To prune kernels and neurons in a CNN, we study the importance levels of neurons based on feature ranking of the final output, and then compute importance scores of intermediate neurons by measuring their effect on the final response.
% Based on our study, we derive the \textit{Neuron Importance Score Propagation} (NISP) algorithm to efficiently propagate final response importance throughout the whole network.

%A naive way would be measuring the importance of neurons/kernels and prune them independently on a layer-by-layer basis, followed by iteratively retraining to compensate the loss of accuracy. However, since the feature dimension is very high and the datasets are usually in large scale, this method becomes computationally intractable. 
An overview of NISP is illustrated in Fig.~\ref{fig:spe}. Given a trained CNN, we first apply a feature ranking algorithm on this final response layer and obtain the importance score of each neuron. Then, the proposed NISP algorithm propagates importance scores throughout the network. 
% The theory and algorithm of RIP is described in Sec.~\ref{sec:rip}. 
Finally, the network is pruned based on the importance scores of neurons and fine-tuned to recover its accuracy.
% The three components are described separately as follows.

%%%%%%%%%%%%%%%%% feature ranking
\subsection{Feature Ranking on the Final Response Layer}\label{sec:featrank}
% The predictive capability of a neural network is highly dependent on the quality of the features in the final layer response. 
% Existing work has demonstrated that a subset of important features, selected based on feature ranking of the network's final response, still retains good predictive power \cite{Roffo_2015_ICCV}.
Our intuition is that the final responses of a neural network should play key roles in full network pruning since they are the direct inputs of the classification task. So, in the first step, we apply feature ranking on the final responses.

It is worth noting that our method can work with any feature selection that scores features \wrt their classification power. We employ the recently introduced filtering method Inf-FS ~\cite{Roffo_2015_ICCV} because of its efficiency and effectiveness on CNN feature selection. 
% Inf-FS first constructs an affinity graph whose nodes are the features to be scored and whose edges represent feature relations like correlation. A selection of features is represented by a path through the graph passing through those features. 
Inf-FS utilizes properties of the power series of matrices to efficiently compute the  importance of a feature with respect to all the other features, \ie, it is able to integrate the importance of a feature over all paths in the affinity graph\footnote{Details of the method are introduced in \cite{Roffo_2015_ICCV} and its codes taken from {\scriptsize\url{https://www.mathworks.com/matlabcentral/fileexchange/54763-infinite-feature-selection-2016}}.}. 

%%%%%%%%%%%%%%%%%%%%%%%%% RIP
\subsection{Neuron Importance Score Propagation (NISP)}\label{sec:rip}
Our goal is to decide which intermediate neurons to delete, given the importance scores of final responses, so that the predictive power of the network is maximally retained. We formulate this problem as a binary integer programming (optimization) and provide a closed-form approximate solution. Based on our theoretical analysis, we develop the \textit{Neuron Importance Score Propagation} algorithm to efficiently compute the neuron importance for the whole network.

%\begin{figure}[h]
%\centering     
%\subfigure[]{\label{fig:proof1}\includegraphics[width=30mm]{image/proof1}}
%\subfigure[]{\label{fig:proof2}\includegraphics[width=30mm]{image/proof2}}
% \caption{Comparisons with layer-by-layer and magnitude based pruning baselines.}
%\caption{(a) is the case of two consecutive layers. (b) is the case that one low-level layer influences two high-level layers. $\beta$ and $\gamma$ are the importance of the high-level layers, and $\alpha$ is an indicator that determines whether a neuron is pruned in the low-level layer. $\mathbf{W}$ and $\mathbf{W'}$ are the weights between two layers. $\sigma$ is the activation function between layers.}
%\label{fig}
%Pruning based on magnitude
%\end{figure}

%To be general, any two layers of a CNN can be represented as Figure \ref{fig:proof1}, where 

\subsubsection{Problem Definition}\label{sec:probdef}
The goal of pruning is to remove neurons while minimizing accuracy loss. Since model accuracy is dependent on the final responses,
we define our objective as minimizing the weighted distance between the original final responses and the final responses after neurons are pruned of a specific layer.  In following, we use bold symbols to represent vectors and matrices.
% We start with a definition of the general layer formulation. 

%%%%%%%%%%%%%%%%%%%%%%%%%%%%%%%%%%%% definition
\iffalse
Let $\mb x$ be the input to a layer. A layer can be represented using the following general form,
\begin{equation}
\label{proof1_objective_2}
f(\mb x)=\sigma(\mb w\mb x+\mathbf b)
\end{equation}
where $\sigma$ is an activation function, $\mb w,\mb b$ are weight and bias. The $j$-th output element of function $f$ is
$f_j(\mb x) = \sigma\left(\sum_{i} w_{i,j}x_i+b_j\right)$
where $x_i$ is the $i$-th element of $\mb x$.
\fi
%\end{definition}
%%%%%%%%%%%%%%%%%%%%%%%%%%%%%%%%%%%%%%%%%

Most neural networks can be represented as a nested function. Thus, we define a network with depth $n$ as a function $F^{(n)}=f^{(n)}\circ f^{(n-1)}\circ\dots\circ f^{(1)}$. The $l$-th layer $f^{(l)}$ is represented using the following general form,
\begin{equation}
\label{proof1_objective_2}
f^{(l)}(\mb x)=\sigma^{(l)}(\mb w^{(l)}\mb x+\mathbf b^{(l)}),
\end{equation}
where $\sigma^{(l)}$ is an activation function and $\mb w^{(l)},\mb b^{(l)}$ are weight and bias, and f(n) represents the "final response layer".
Networks with branch connections such as the skip connection in ResNet can be transformed to this representation by padding weights and merging layers. % with added binary (zero or one) weights.

%The $j$-th output element of function $f^{(k)}$ is $f_j(\mb x) = \sigma\left(\sum_{i} w_{i,j}x_i+b_j\right)$ where $x_i$ is the $i$-th element of $\mb x$.

% \begin{definition}[Neuron prune indicator]\label{def:indicator}
We define the \textit{neuron importance score} as a non-negative value \wrt a neuron, and use $\mb s_l$ to represent the vector of neuron importance scores  in the $l$-th layer. Suppose $N_l$ neurons are to be kept in the $l$-th layer after pruning; we define the \textit{neuron prune indicator} of the $l$-th layer as a binary vector $\mb s^*_l$, computed based on neuron importance scores $\mb s_l$ such that $s^*_{l,i}=1$ if and only if $s_{l,i}$ is among top $N_l$ values in $\mb s_l$. 

%This definition implies that given all neuron prune indicators $\{\mb s^*_1,\mb s^*_2,\ldots,\mb s^*_{N}\}$ where $\mb s^*_k$ is the indicator for keeping $k$ neurons after pruning, any non-negative vector $\mb s$ fulfilling such condition that $s_{i}\le s_{j}\Rightarrow s^*_{n,i}\le s^*_{n,j},\forall i,j,n$ is a feasible neuron importance score vector.
% \end{definition}

% \begin{proposition}[Neuron importance score]

% \begin{proof}
% Implied from Definition~\ref{def:indicator}.
% \end{proof}
% \end{proposition}

\begin{figure}[t]
\begin{center}
   \includegraphics[width=0.75\linewidth]{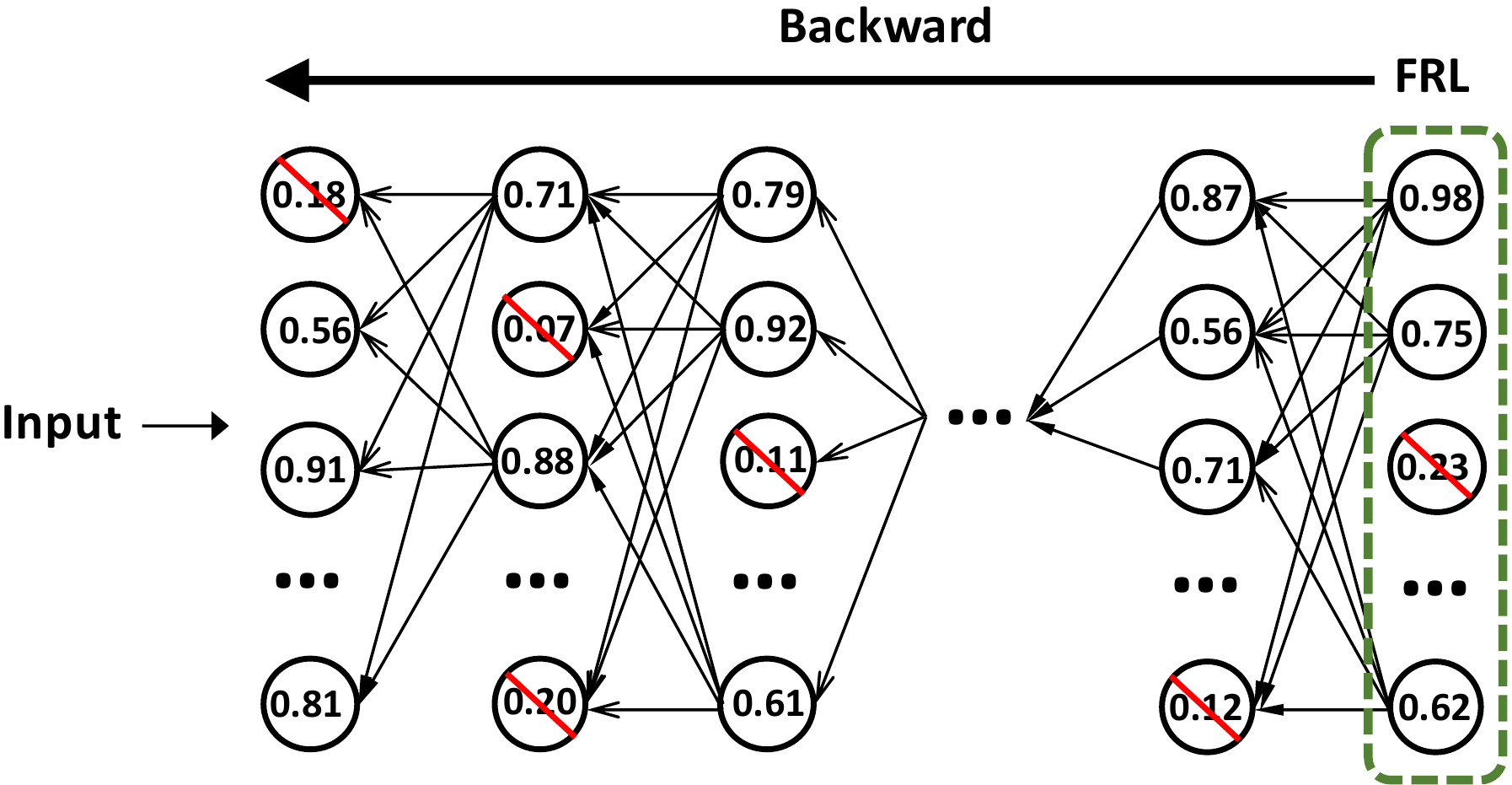}
\end{center}
   \caption{We propagate the neuron importance from the final response layer (FRL) to previous layers, and prune bottom-ranked neurons (with low importance scores shown in each node) given a pre-defined pruning ratio per layer in a single pass. The importance of pruned neurons (with backslash) is not propagated.} 
\label{fig:NISP}
\end{figure}

%%%%%%%%%%%%%%%%%%%%%%%%%%%%%%%%%%% objective function
\subsubsection{Objective Function}  The motivation of our objective is that the difference between the responses produced by the original network and the one produced by the pruned network should be minimized w.r.t. important neurons. Let $F^{(n)}$ be a neural network with $n$ layers. Suppose we have a dataset of $M$ samples, and each is represented using $\mb x^{(m)}_0$. For the $m$-th sample, we use $\mb x^{(m)}_l$ to represent the response of the $l$-th layer (which is the input to the $(l+1)$-th layer). The final output of the network is $\mb x^{(m)}_n$ and its corresponding non-negative neuron importance is $\mb s_n$. We define \begin{align}\label{eq:gnet}
G^{(i,j)}=f^{(j)}\circ f^{(j-1)}\circ\cdots\circ f^{(i)}
\end{align} as a sub-network of $F^{(n)}$ starting from the $i$-th layer to the $j$-th layer. Our goal is to compute for the $l$-th layer the neuron prune indicator $\mb s^*_l$ so that the influence of pruning the $l$-th layer on the important neurons of the final response is minimized. To accomplish this, we define an optimization objective w.r.t. the $l$-th layer neuron prune indicator, \ie,
\begin{equation}
\label{proof1_objective}
\arg\min_{\mb s^*_l}\ \sum_{m=1}^M \mathcal{F}(\mb s^*_l|\mb x^{(m)}_l,\mb s_n; G^{(l+1,n)})~,
\end{equation}
which is accumulated over all samples in the dataset. The objective function for a single sample is defined as
\begin{equation}\label{proof1_objective2}
\mathcal{F}(\mb s^*_l|\mb x,\mb s_n; F)=\left\langle\ \mb s_n,\ |F(\mb x)-F(\mb s^*_l \odot \mb x)|\ \right\rangle,
\end{equation}
where $\langle\cdot,\cdot\rangle$ is dot product, $\odot$ is element-wise product and $|\cdot|$ is element-wise absolute value. The solution to Eq. \ref{proof1_objective} indicates which neurons should be pruned in an arbitrary layer.

%Eq.~\ref{proof1_objective2} can be re-written as
%\begin{align}\label{proof1_FSx}
%\mathcal{F}(\mb s^*_l|\mb x,\mb s_n; F)=\sum_j s_{nj}|F_j(\mb x) - F_j(\mb s^*_l\odot \mb x)|
%\end{align}
%where $s_{nj}$ is the $j$-th element of $\mb s_n$ and $F_j$ is the $j$-th output element of function $F$.

\subsubsection{Solution}
The network pruning problem can be formulated as a binary integer program, finding the optimal neuron prune indicator in Eq.~\ref{proof1_objective}. However, it is hard to obtain efficient analytical solutions by directly optimizing Eq.~\ref{proof1_objective}. So we derive an upper bound on this objective, and show that a sub-optimal solution can be obtained by minimizing the upper bound. Interestingly, we find a feasible and efficient formulation for the importance scores of all neurons based on this sub-optimal solution. 

Recall that the $k$-th layer is defined as $f^{(k)}(\mb x)=\sigma^{(k)}(\mb w^{(k)}\mb x+\mb b^{(k)})$. We assume the activation function $\sigma^{(k)}$ is Lipschitz continuous since it is generally true for most of the commonly used activations in neural networks  such as Identity, ReLU, sigmoid, tanh, PReLU, \etc. Then we know for any $\mb x,\mb y$, there exists a constant $C_\sigma^{(k)}$ such that $|\sigma^{(k)}(\mb x)-\sigma^{(k)}(\mb y)|\le C_\sigma^{(k)}|\mb x-\mb y|$.
%\begin{align}
%    |\sigma^{(k)}(\mb x)-\sigma^{(k)}(\mb y)|\le C_\sigma^{(k)}|\mb x-\mb y|~.
%\end{align}
Then it is easy to see
\begin{align}\label{lemma2}
|f^{(k)}(\mb x)-f^{(k)}(\mb y)|\le C_\sigma^{(k)}|\mb w^{(k)}|\cdot|\mb x-\mb y|~,
\end{align}
where $|\cdot|$ is the element-wise absolute value. From Eq.~\ref{eq:gnet}, we see that $G^{(i,j)}=f^{(j)}\circ G^{(i,j-1)}$. Therefore, we have,
\begin{align}
    &|G^{(i,j)}(\mb x)-G^{(i,j)}(\mb y)|\nonumber\\
    &~~~~~~~\le C_\sigma^{(j)}|\mb w^{(j)}||G^{(i,j-1)}(\mb x)-G^{(i,j-1)}(\mb y)|~.\label{eq:G}
\end{align}
Applying Eq.~\ref{lemma2} and Eq.~\ref{eq:G} repeatedly, we have, $\forall i\le j\le n$,
\begin{align}
   & |G^{(i,n)}(\mb x)-G^{(i,n)}(\mb y)|\le C_\Sigma^{(i,n)}\mb W^{(i,n)}|\mb x-\mb y|,\label{eq:intermediate}
\end{align}
where $\mb W^{(i,j)}=|\mb w^{(j)}||\mb w^{(j-1)}|\cdots|\mb w^{(i)}|$, and  $C_\Sigma^{(i,j)}=\prod_{k=i}^j C_\sigma^{(k)}$.
Substituting $\mb x = \mb x_l^{(m)},\mb y=\mb s^*_l\odot\mb x_l^{(m)},i=l+1$ into Eq.~\ref{eq:intermediate}, we have
\begin{align}
&|G^{(l+1,n)}(\mb x^{(m)}_l)-G^{(l+1,n)}(\mb s^*_l\odot\mb x^{(m)}_l)|\nonumber\\
&~~~~~~\le C_\Sigma^{(l+1,n)}\mb W^{(l+1,n)}|\mb x^{(m)}_l-\mb s^*_l\odot\mb x^{(m)}_l|~.
\end{align}
Since $\mb s_n$ is a non-negative vector,
\begin{align}
&\mathcal{F}(\mb s^*_l|\mb x^{(m)}_l,\mb s_n; G^{(l+1,n)})\nonumber\\
&~~~~=\langle\mb s_n, |G^{(l+1,n)}(\mb x^{(m)}_l)-G^{(l+1,n)}(\mb s^*_l\odot\mb x^{(m)}_l)|\rangle\\
&~~~~\le \langle\mb s_n, C_\Sigma^{(l+1,n)}\mb W^{(l+1,n)}|\mb x^{(m)}_l-\mb s^*_l\odot\mb x^{(m)}_l|\rangle\\
&~~~~=C_\Sigma^{(l+1,n)}\langle{\mb W^{(l+1,n)}}^\intercal\mb s_n, (\mb 1-\mb s^*_l)\odot  |\mb{x}^{(m)}_l|\rangle~.
\end{align}
Let us define $\mb r_l={\mb W^{(l+1,n)}}^\intercal\mb s_n$; then
\begin{align}
&\textstyle\sum_{m=1}^M \mathcal{F}(\mb s^*_l|\mb x^{(m)}_l,\mb s_n; G^{(l+1,n)})\nonumber\\
&\textstyle~~~~~~\le C_\Sigma^{(l+1,n)}\sum_{m=1}^M \langle\mb r_l,(\mb 1-\mb s^*_l)\odot|\mb x^{(m)}_l|\rangle\\
&\textstyle~~~~~~\le C_\Sigma^{(l+1,n)}\sum_{m=1}^M \sum_i r_{l,i}(1-s^*_{l,i})|x^{(m)}_{l,i}|\\
&\textstyle~~~~~~=C_\Sigma^{(l+1,n)}\sum_ir_{l,i}(1-s^*_{l,i})\sum_{m=1}^M|x^{(m)}_{l,i}|~.
\end{align}
Since $|\mb x^{(m)}_{l,i}|$ is bounded, there must exist a constant $C_x$ such that $\sum_{m=1}^M |x^{(m)}_{l,i}|\le C_x,\forall i$. Thus, we have
\begin{align}
&\sum_{m=1}^M \mathcal{F}(\mb s^*_l|\mb x^{(m)}_l,\mb s_n; F^{(l+1)})\le C\sum_ir_{l,i}(1-s^*_{l,i}),\label{eq:upbound}
\end{align}
where $C=C_\Sigma^{(l+1,n)}C_x$ is a constant factor.

Eq.~\ref{eq:upbound} reveals an upper-bound of our objective in Eq.~\ref{proof1_objective}. Thus, we minimize this upper-bound, \ie,
\begin{align}
\arg\min_{\mb s^*_l}\sum_ir_{l,i}(1-s^*_{l,i})\Leftrightarrow\arg\max_{\mb s^*_l}\ \sum_is^*_{l,i}r_{l,i}~.\label{suboptimal}
\end{align}
%So it is equivalent to
%\begin{align}\label{theory:finalobjective}
%\arg\max_{\mb s^*_x}\ \sum_is^*_{xi}r_i~.
%\end{align}
The optimal solution to Eq.\ref{suboptimal} is sub-optimal with respect to the original objective in Eq.~\ref{proof1_objective}, however it still captures the importance of neurons. It is easy to see that if we keep $N_x$ neurons in the $l$-th layer after pruning, then the solution to Eq.~\ref{suboptimal} is that $s^*_{l,i}=1$ if and only if $r_{l,i}$ is among the highest $N_x$ values in $\mb r_l$. According to the definition of neuron prune indicator in Sec.~\ref{sec:probdef}, $\mb r_l={\mb W^{(l+1,n)}}^\intercal\mb s_n$ is a feasible solution to the importance scores of the $l$-th layer response. This conclusion can be applied to every layer in the network. Based on this result, we define the neuron importance of a network as follows.
 %We can see, for any $N_x$, $r_i\le r_j\Rightarrow s^*_{xi}\le x^*_{xj}$. 

\begin{definition}[Neuron importance score]\label{def:ri}
Given a neural network $F^{(n)}$ containing $n$ layers and the importance score  $\mb s^{(n)}$ of the last layer response, the importance score of the $k$-th layer response can be computed as 
\begin{align}
\mb s_k=|\mb w^{(k+1)}|^\intercal|\mb w^{(k+2)}|^\intercal\cdots|\mb w^{(n)}|^\intercal \mb s_n,
\end{align}
where $\mb w^{(i)}$ is the weight matrix of the $i$-th layer.
\end{definition}

An important property of neuron importance is that it can be computed recursively (or propagated) along the network.
\begin{proposition}[Neuron importance score propagation] \label{def:rip} The importance score of the $k^\text{th}$ layer response can be propagated from the importance score of the $(k+1)^\text{th}$ layer by
\begin{align}
\mb s_k=|\mb w^{(k+1)}|^\intercal\mb s_{k+1},\label{eq:rip}
\end{align}
where $\mb w^{(k+1)}$ is the weight matrix of the $(k+1)^\text{th}$ layer.
\end{proposition}

%Corollary \ref{corollary:rip} reveals that the importance of neurons in a low-level layer activation can be propagated from its direct higher-level layer activation by the magnitude of weights of the layer in between.

\subsubsection{Algorithm}
We propose the \textit{Neuron Importance Score Propagation} (NISP) algorithm (shown in Fig. \ref{fig:NISP}) based on Proposition~\ref{def:rip}. Initially, we have the importance score of every neuron in the final response layer of the network. Definition~\ref{def:ri} shows that the importance score of every other layer in the network is directly correlated with the importance of the final response. However, instead of computing the importance expensively using Definition~\ref{def:ri}, we see from Eq.~\ref {eq:rip} that the importance score of a lower layer can be propagated directly from the adjacent layer above it. An equivalent form of Eq.~\ref{eq:rip} is
\begin{equation}
    \textstyle s_{k,j}=\sum_i |w^{(k+1)}_{i,j}|s_{k+1,i},\label{eq:prop}
\end{equation}
where $s_{k,j}$ is the importance score of the $j$-th neuron in the $k$-th layer response. 

We conclude from Eq.~\ref{eq:prop} that the importance of a neuron is a weighted sum of all the subsequent neurons that are directly connected to it. This conclusion also applies to normalization, pooling and branch connections in the network (\ie, a layer is directly connected with multiple layers)\footnote{\label{supp}See supplementary material for more details and proofs.}. The NISP algorithm starts with the importance in FRL and repeats the propagation (Eq.~\ref{eq:prop}) to obtain the importance of all neurons in the network with a single backward pass (Fig.~\ref{fig:spe}).

%%% @todo(ang): Put this into supplementary materials.
%The importance of batch normalization layers is propagated identically since the connections between neurons form a one-one mapping. An exception is the max-pooling layer in which we assume a uniform probability for each neuron to be the maximum in the window. So in our implementation, the NISP of max-pooling is treated in the same way as average-pooling. Our algorithm starts with the final layer importance and repeats the propagation (Eq.~\ref{eq:prop}) to obtain the importance of all neurons in the network with a single backward pass (Fig.~\ref{fig:spe}).
% \footref{supp}. 

%RIP efficiently measures the importance of neurons and kernels of an entire deep neural network consistently as shown in Figure \ref{fig:spe}. Importantly, it is able to assign credit to neurons that have important contributions to future layers but which might otherwise be scored as unimportant if evaluated as an independent feature.

%The importance of a neuron in a low-level layer is computed based on the importance of its next layer's neurons and the weights between them as follows:
%\begin{equation}
%\label{RIP}
%s^{(k)}_j=\sum_i s^{(k+1)}_j |w^{(k)}_{i,j}|
%\end{equation}
%where $n_i$ is the $i^{th}$ neuron in the %low-level layer, $\beta_j$ is the importance of %the $j^{th}$ neuron in the next layer and %$w_{i,j}$ is the weight between them. We propagate %the importance from the final response layer to %the lowest layer of a CNN layer-by-layer using %Eq.~\ref{RIP}.

\subsection{Pruning Networks Using NISP} \label{sec:prune} Given target pruning ratios for each layer, we propagate the importance scores, compute the prune indicator of neurons based on their importance scores 
% (Definition~\ref{def:indicator}) 
and remove neurons with prune indicator value $0$. The importance propagation and layer pruning happens jointly in a single backward pass, and the importance of a pruned neuron is not propagated to any further low-level layers. For fully connected layers, we prune each individual neuron. For convolution layers, we prune a whole channel of neurons together. The importance score of a channel is computed as the summation of the importance scores of all neurons within this channel\footref{supp}.%\footnote{More details can be found in supplementary materials.}.

\section{Experiments}
We evaluate our approach on standard datasets with popular CNN networks. We first compare to \textit{random pruning} and \textit{training-from-scratch} baselines to demonstrate the effectiveness of our method. We then compare to two other baselines, \textit{magnitude-based pruning} and \textit{layer-by-layer pruning} to highlight the contributions of feature ranking and neuron importance score propagation, respectively.
% %, which measures the importance of neurons in the final response layer based on the absolute sum of the weights between neurons in the previous layer and the final layer, then propagate the importance using RIP; layer-by-layer pruning, which conducts feature ranking and prune neurons for each layer independently. 
%A comparison on different pruning ratios is presented afterwards. 
Finally, we benchmark the pruning results and compare to existing methods such as \cite{PerforatedCNN,Tucker,learning,pruneweigth}.

\subsection{Experimental Setting}\label{sec:expset}

%\textbf{Datasets.} 
We conduct experiments on three datasets, MNIST~\cite{lenet}, CIFAR10 and ImageNet~\cite{imagenet_cvpr09}, for the image classification task.
%\textbf{Architectures.} 
We evaluate using five commonly used CNN architectures: \textit{LeNet} \cite{lenet}, \textit{Cifar-net}\footnote{\scriptsize\url{https://code.google.com/p/cuda-convnet/}.}, \textit{AlexNet} \cite{Alexnet}, \textit{GoogLeNet} \cite{googlenet} and \textit{ResNet} \cite{resnet}.

All experiments and time benchmarks are obtained using Caffe ~\cite{caffe}.
The hyper-parameter of Inf-FS is a loading coefficient $\alpha \in [0, 1] $, which controls the influence of variance and correlation when measuring the importance. 
We conduct PCA accumulated energy analysis (results shown in the supplementary material) as suggested in \cite{Nonlinear} to guide our choice of pruning ratios. 
% We choose the default pruning ratio for each layer as 50\% to evaluate our method. We also provide results to demonstrate the effective of our method with different pruning ratios.
% Examples of PCA accumulated energy analysis and the experimental results that demonstrate the effectiveness of our method with different pruning ratios can be found in the supplementary materials. 

\begin{figure*}[!t]
\centering     %%% not \center
\subfigure[MNIST]{\label{fig:Mnist}\includegraphics[height=3.5cm,width=.24\linewidth]{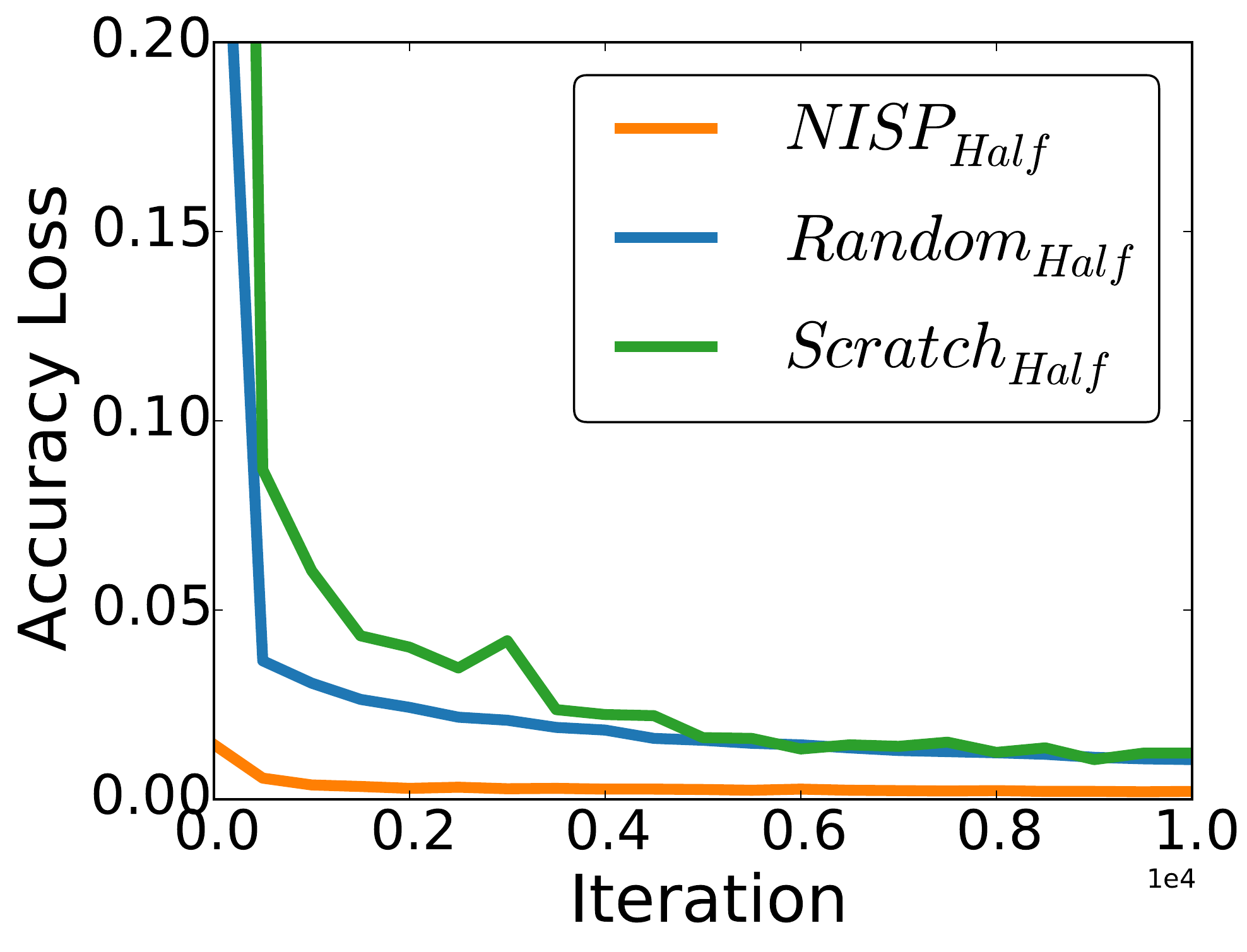}}
\subfigure[CIFAR10]{\label{fig:cifar}\includegraphics[height=3.5cm,width=.24\linewidth]{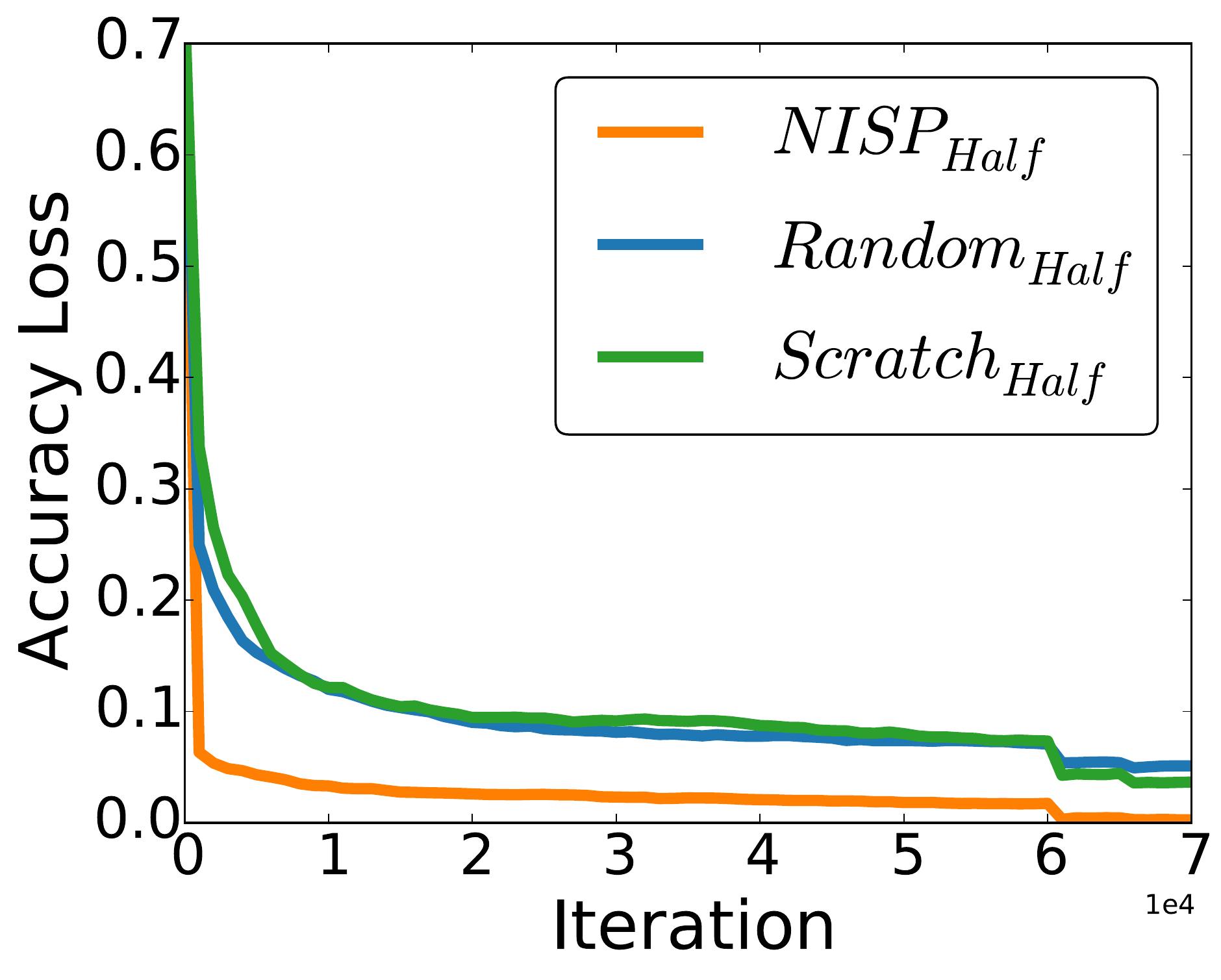}}
\subfigure[ImageNet: AlexNet]{\label{fig:alexbaseline}\includegraphics[height=3.5cm,width=.24\linewidth]{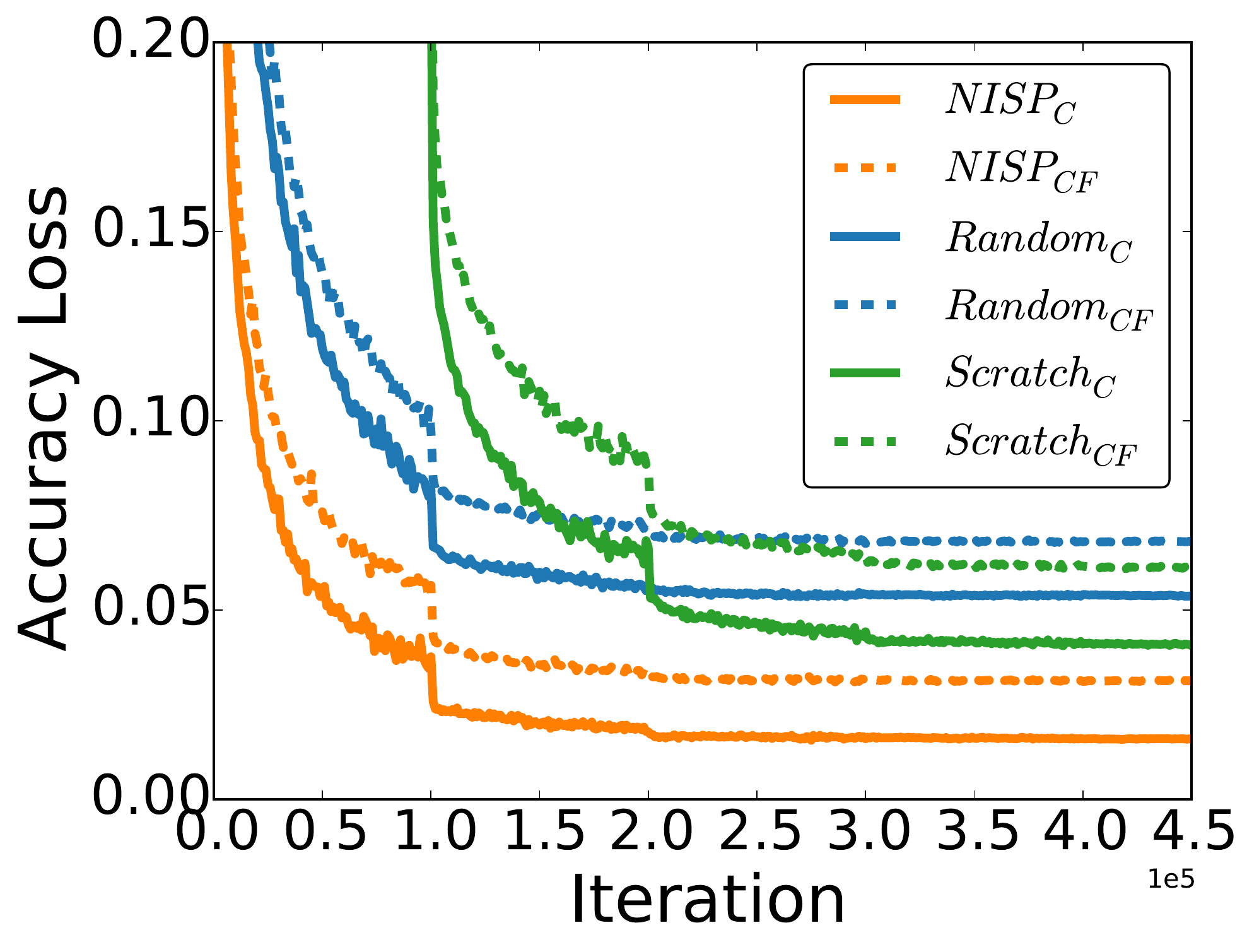}}
\subfigure[ImageNet: GoogeLeNet]{\label{fig:google}\includegraphics[height=3.5cm,width=.24\linewidth]{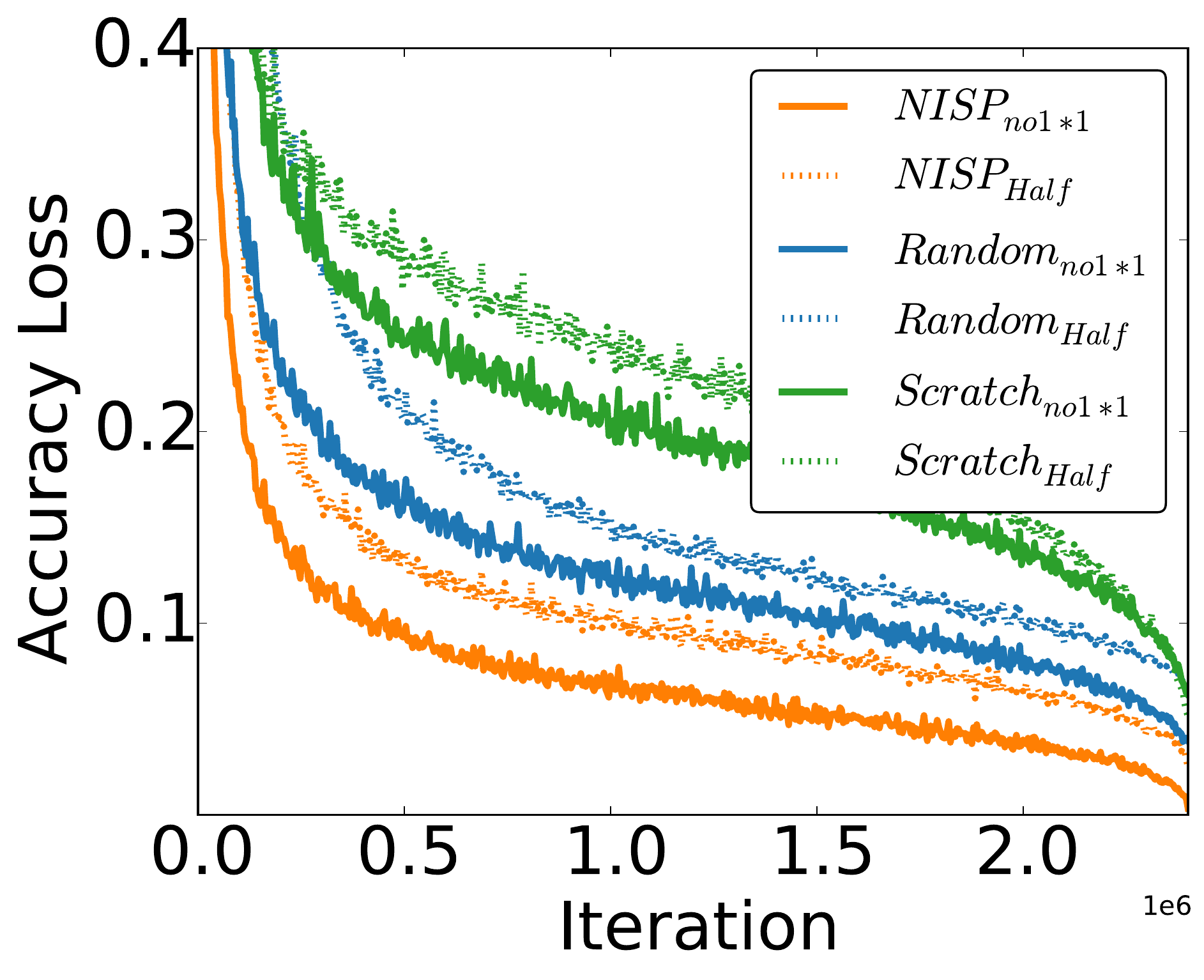}}
\caption{Learning curves of random pruning and training from scratch baselines and NISP using different CNNs on different datasets. The pruning ratio of neurons and filters is 50\%. Networks pruned by NISP (orange curves) converge the fastest with the lowest accuracy loss.}\label{fig:cifar10}
\end{figure*}

\subsection{Comparison with Random Pruning and Train-from-scratch Baselines}\label{sec:naiveexp}
We compare to two baselines: (1) randomly pruning the pre-trained CNN and then fine-tuning, and (2) training a small CNN with the same number of neurons/filters per layer as our pruned model from scratch. 
We use the same experimental settings for our method and baselines except for the initial learning rate.
For training from scratch, we set the initial learning rate to the original one, while for fine-tuning tasks (both NISP and random pruning), the initial learning rate is reduced by a factor of 10. 

\textbf{LeNet on MNIST:}
We prune half of the neurons in FC layers and half of the filters in both convolution layers in Fig. \ref{fig:Mnist}. Our method is denoted as $\textsl{NISP}_\textsl{Half}$, while the baseline methods that prune randomly or train from scratch are denoted as $\textsl{Random}_\textsl{Half}$ and $\textsl{Scratch}_\textsl{Half}$. 
Our method outperforms the baselines in three aspects.
First, for fine-tuning (after pruning), unlike the baselines, our method has very small accuracy loss at iteration 0; this implies that it retains the most important neurons, pruning only redundant or less discriminative ones.
Second, our method converges much faster than the baselines.
Third, our method has the smallest accuracy loss after fine-tuning. 
For LeNet on MNIST, our method only decreases 0.02\% top-1 accuracy with a pruning ratio of 50\% as compared to the pre-pruned network.

\textbf{Cifar-net on CIFAR10:}
The learning curves are shown in Fig.~\ref{fig:cifar}. 
Similar to the observations from the experiment for LeNet on MNIST, our method outperforms the baselines in the same three aspects: the lowest initial loss of accuracy, the highest convergence speed and the lowest accuracy loss after fine-tuning. Our method has less than 1\% top-1 accuracy loss with 50\% pruning ratio for each layer.

\textbf{AlexNet on ImageNet:}
To demonstrate that our method works on large and deep CNNs, we replicate experiments on AlexNet with a pruning ratio of 50\% for all convolution layers and FC layers (denoted as {$\textsl{NISP}_\textsl{CF}$} when we prune both conv and FC layers). 
Considering the importance of FC layers in AlexNet, we compare one more scenario in which our approach only prunes half of the filters but without pruning neurons in FC layers (denoted as $\textsl{NISP}_\textsl{C}$).
% We use the same hyper-parameters used in training AlexNet in our experiment, and reduce the initial learning rate by a factor of 10 for fine-tuning. 
We reduce the initial learning rate by a factor of 10, then fine-tune 90 epochs and report top-5 accuracy loss.
%The learning curves are shown in Fig. \ref{fig:alexbaseline}.
Fig. \ref{fig:alexbaseline} shows that for both cases (pruning both convolution and FC layers and pruning only convolution layers), the advantages we observed on MNIST and CIFAR10 still hold. 
Layer-wise computational reduction analysis that shows the full-network acceleration can be found in supplementary materials.

\textbf{GoogLeNet on ImageNet:}
We denote the reduction layers in an inception module as ``Reduce", and the $1 \times 1$ convolution layer without reduction as ``1$\times$1".
We use the quick solver from Caffe in training. 
% For fine-tuning, we use an initial learning rate of 0.001. We use the same platforms as in section \ref{existing} in this experiment.
% The batch size for time benchmark is 1 for fair comparison with \cite{Tucker}. 
We conduct experiments between our method and the baselines for 3 pruning strategies: (\textsl{Half}) pruning all convolution layers by half; (\textsl{noReduce}) pruning every convolution layer except for the reduction layers in inception modules by half; (\textsl{no1x1}) pruning every convolution layer  by half except the $1 \times 1$ layers in inception modules. We show results for two of them in Fig. \ref{fig:google}, and observe similar patterns to the experiments on other CNN networks\footnote{See supplementary materials for the results of \textsl{noReduce}.}. For all GoogLeNet experiments, we train/fine-tune for 60 epochs and report top-5 accuracy loss.

% \subsection{Ablation Study}
% Our method has two merits: (1) it measures the importance of neurons by feature ranking, which jointly considers both CNN weights and input data, and (2) it measures the importance of all neurons globally considering inter-layers connections.

\subsection{Feature Selection v.s. Magnitude of Weights}
How to define neuron importance is an open problem. Besides using feature ranking to measure neuron importance, other methods \cite{pruneweigth,thinet,DeepCompress} measure neuron importance by magnitude of weights. To study the effects of different criteria to determine neuron importance, we conduct experiments by fixing other parts of NISP and only comparing the pruning results with different measurements of importance: 1. using feature selection method in \cite{Roffo_2015_ICCV} (NISP-FS) and 2. considering only magnitude of weights (NISP-Mag).
For the Magnitude-based pruning, the importance of a neuron in the final response layer equals the absolute sum of all weights connecting the neuron with its previous layer. To compare only the two metrics of importance, we rank the importance of neurons in the final response layer based on the magnitude of their weight values, and propagate their importance to the lower layers. Finally, we prune and fine-tune the model in the same way as the NISP method. 

For the ``NISP-Mag" baseline, we use both AlexNet and Cifar-net architectures. The learning curves of those baselines are shown in Fig. \ref{fig:maglbl}. We observe that ``NISP-FS" yields much smaller accuracy loss with the same pruning ratio than ``NISP-Mag", but ``NISP-Mag" still outperforms the random pruning and train-from-scratch baselines, which shows the effectiveness of NISP with different measurement of importance. In the remainder of this paper, we employ the feature ranking method proposed in \cite{Roffo_2015_ICCV} in NISP.

% Meanwhile, the method in \cite{pruneweigth} is also based on the magnitude of weight values. We show that our method outperforms \cite{pruneweigth} in the following section.

\subsection{NISP v.s. Layer-by-Layer Pruning}\label{LBL}
To demonstrate the advantage of the NISP's importance propagation, we compare with a pruning method that conducts feature ranking on every layer to measure the neuron importance and prune the unimportant neurons of each layer independently. All other settings are the same as NISP. We call this method ``Layer-by-Layer"  (LbL) pruning.
  
\begin{figure}[!t]
\centering    
\subfigure[AlexNet on ImageNet]{\label{fig:mag}\includegraphics[height=3.5cm,width=.49\linewidth]{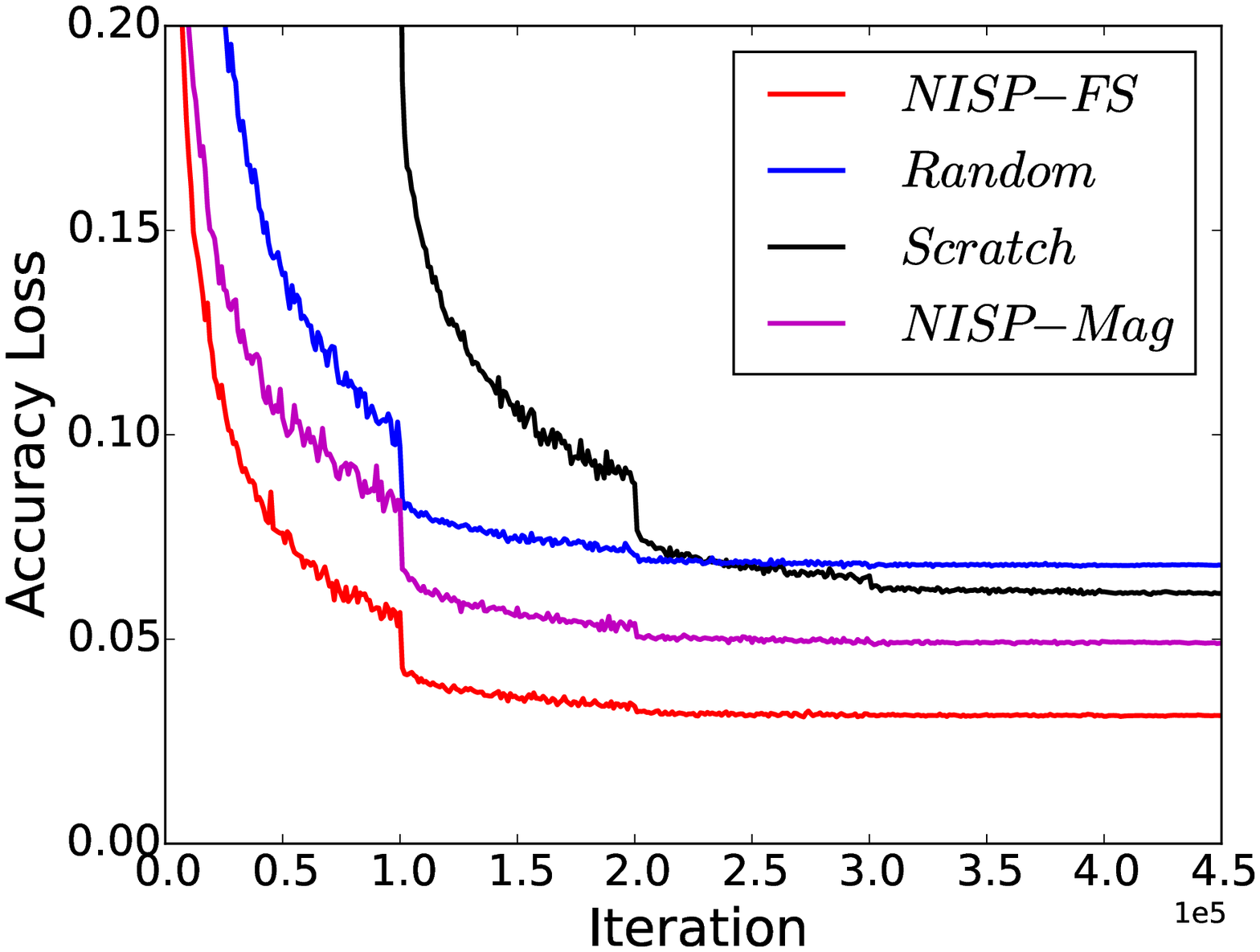}}
\subfigure[Cifar-net on CIFAR10]{\label{fig:layer}\includegraphics[height=3.5cm,width=.49\linewidth]{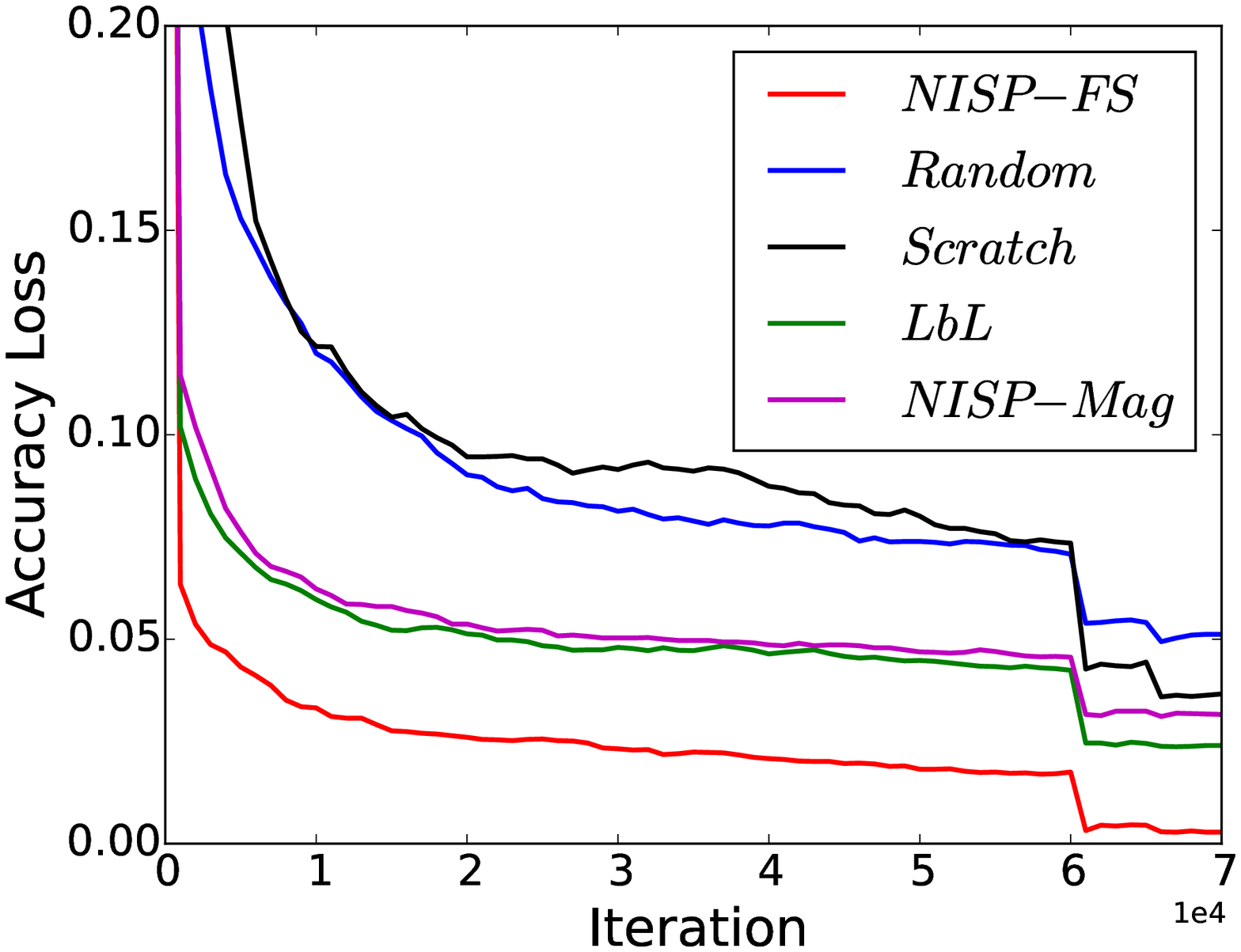}}
% \caption{Comparisons with layer-by-layer and magnitude based pruning baselines.}
\caption{Comparison with layer-by-layer (LbL) and magnitude based (Mag) pruning baselines. We prune 50\% of neurons and filters in all layers for both CNNs. NISP-FS outperforms NISP-Mag and LbL in terms of prediction accuracy.}
\label{fig:maglbl}
%Pruning based on magnitude
\end{figure}

One challenge for the ``LbL" baseline is that the computational cost of measuring neuron importance on each layer is huge. So we choose a small CNN structure trained on the CIFAR10 dataset. 
Fig. \ref{fig:layer} shows that although the ``LbL" method outperforms the baselines, it performs much worse than NISP in terms of the final accuracy loss with the same pruning ratio, which shows the need for measuring the neuron importance across the entire network using NISP.

To further study the advantage of NISP over layer-by-layer pruning, we define the Weighted Average Reconstruction Error (WARE) to measure the change of the important neurons' responses on the final response layer after pruning (without fine-tuning) as:
\begin{equation}
\text{WARE} = \frac{\sum_{m=1}^{M} \sum_{i=1}^{N} s_{i}\cdot \frac{|{\hat y}_{i,m}- y_{i,m}|}{|y_{i,m}|}}{M \cdot N},
\end{equation}
where $M$ and $N$ are the number of samples and number of retained neurons in the final response layer; $s_i$ is the importance score; $y_{i,m}$ and ${\hat y}_{i,m}$ is the response on the $m^{th}$ sample of the $i^{th}$ neuron before/after pruning. 

\begin{figure}[t]
\begin{center}
   \includegraphics[width=0.65\linewidth]{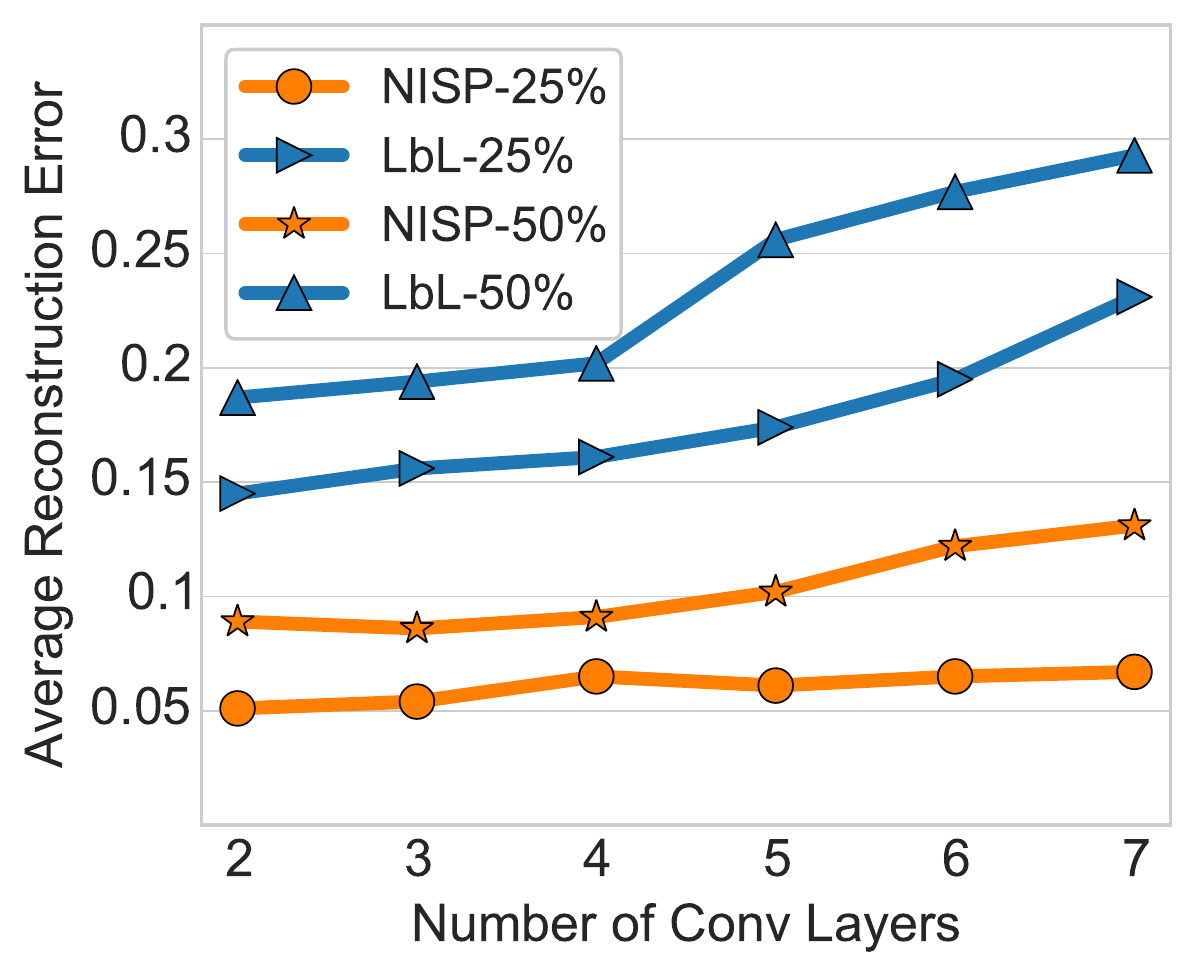}
\end{center}
   \caption{Weighted Average Reconstruction Error (WARE) on the final responses without fine-tuning: we set pruning ratios as 25\% and 50\% and evaluate the WARE on the final responses of models with different depths pruned using NISP or LbL. It is clear that networks pruned by NISP have the lowest reconstruction errors.}
\label{fig:LBL}
\end{figure}

We design different Cifar-net-like CNNs with different numbers of Conv layers, and apply NISP and LbL pruning with different pruning ratios. We report the WARE on the retained neurons in the final response layer (``ip1" layer in Cifar-net-like CNNs) in Fig. \ref{fig:LBL}. We observe that: 1. As network depth increases, the WARE of the LbL-pruned network dramatically increases, which indicates the error propagation problem of layer-by-layer pruning, especially when the network is deep, and suggests the need for a global pruning method such as NISP; 2. The WARE of the LbL method becomes much larger when the pruning ratio is large, but is more stable when using NISP to prune a network; 3. NISP methods always reduce WARE on the retained neurons compared to LbL. The small reconstruction errors on the important neurons in the final response layer obtained by NISP provides a better initialization for fine-tuning, which leads to much lower accuracy loss of the pruned network.

\begin{table}[h]
\centering
\setlength{\tabcolsep}{4pt} 
\footnotesize
\begin{tabular}{@{}llccc@{}}
\toprule
                        & Model                           &
                        Accu.$\downarrow$\% & FLOPs$\downarrow$\% & Params.$\downarrow$\% \\ \midrule
\multirow{1}{*}{AlexNet }                 & NISP-A                          & \textbf{1.43}                          & \textbf{67.85}                         & 33.77                                        \\
\multirow{1}{*}{on ImageNet }& Perforated \cite{PerforatedCNN}         & 2.00                          & 50.00                         & -                                         \\\cmidrule{2-5}
                        & NISP-B                       & \textbf{0.97}                          & \textbf{62.69}                         & 1.96                                       \\
                        & Tucker \cite{Tucker}                & 1.70                          & 62.55                         & -                                        \\\cmidrule{2-5}
                        & NISP-C                    & \textbf{0.54}                          & \textbf{53.70}                         & 2.91                                      \\
                        & Learning \cite{learning} &1.20                          & 48.19                         & -                                          \\
                        \cmidrule{2-5}
                        & NISP-D                     & 0.00                           & 40.12                         & 47.09                             \\\midrule
\multirow{1}{*}{GoogLeNet} & NISP  & \textbf{0.21}                          & \bf{58.34}                        & \textbf{33.76}                                           \\
 \multirow{1}{*}{on ImageNet }         & Tucker \cite{Tucker} & 0.24    
& 51.50                         & 31.88                                                  \\\midrule

\multirow{1}{*}{ResNet}& NISP-56        & 0.03                         & \textbf{43.61}                         & \textbf{42.60}                                                  \\
 \multirow{1}{*}{on CIFAR10 } & 56-A \cite{pruneweigth}  & \textbf{-0.06}\tablefootnote{A negative value here indicates an improved model accuracy.}  
& 10.40                         & 9.40                                                \\
  & 56-B \cite{pruneweigth} & -0.02                          & 27.60                         & 13.70                                               \\
                        
                        \cmidrule{2-5}
                        & NISP-110       & 0.18                         & \textbf{43.78}                         & \textbf{43.25}                                                 \\
                        & 110-A \cite{pruneweigth} & \textbf{0.02}                          & 15.90                         & 2.30                                                  \\
                        
                        & 110-B \cite{pruneweigth} & 0.23                          & 38.60                         & 32.40                                                 \\
                        \midrule

\multirow{1}{*}{ResNet} & \multirow{1}{*}{NISP-34-A}  & \multirow{1}{*}{\textbf{0.28}}    &  \multirow{1}{*}{27.32}    & \multirow{1}{*}{27.14}           \\
\multirow{1}{*}{on ImageNet} & \multirow{1}{*}{NISP-34-B}  & \multirow{1}{*}{0.92}    &  \multirow{1}{*}{\textbf{43.76}}    & \multirow{1}{*}{\textbf{43.68}}      
\\ 
& \multirow{1}{*}{Res34 \cite{pruneweigth}}  & \multirow{1}{*}{1.06}    &  \multirow{1}{*}{24.20}    & \multirow{1}{*}{-}       
\\\cmidrule{2-5}

 & \multirow{1}{*}{NISP-50-A}  & \multirow{1}{*}{\textbf{0.21}}    &  \multirow{1}{*}{27.31}    & \multirow{1}{*}{27.12}          \\
& \multirow{1}{*}{NISP-50-B}  & \multirow{1}{*}{0.89}    &  \multirow{1}{*}{\textbf{44.01}}    & \multirow{1}{*}{\textbf{43.82}}        \\

& \multirow{1}{*}{Res50 \cite{thinet}}  & \multirow{1}{*}{0.84}    &  \multirow{1}{*}{36.79}    & \multirow{1}{*}{33.67}      
\\

                        \bottomrule
\end{tabular}
\caption{Compression Benchmark. [Accu.$\downarrow$\%] denotes the absolute accuracy loss; [FLOPs$\downarrow$\%] denotes the reduction of computations; [Params.$\downarrow$\%] demotes the reduction of parameter numbers; }
\label{table:others}
\vspace{3pt}
\end{table}

\subsection{Comparison with Existing Methods}\label{existing}
We compare our method with existing pruning methods on AlexNet, GoogLeNet and ResNet, and show results in Table \ref{table:others}.

We show benchmarks of several pruning strategies in Table \ref{table:others}, and provide additional results in the supplementary materials.
In Table \ref{table:others}, for AlexNet, the pruning ratio is 50\%. NISP-A denotes pruning all Conv layers; NISP-B denotes pruning all Conv layers except for Conv5; NISP-C denotes pruning all Conv layers except for Conv5 and Conv4; NISP-D means pruning Conv2, Conv3 and FC6 layers.
For GoogLeNet, we use the similar the pruning ratios of the 3$\times$3 layers in \cite{Tucker}, and we prune 20\% of the
reduce layers. Our method is denoted as ``NISP".

To compare theoretical speedup, we report reduction in the number of multiplication and the number of parameters following \cite{Tucker} and \cite{PerforatedCNN}, and denote them as [FLOPs$\downarrow$$\%$] and [Params.$\downarrow$$\%$] in the table. Pruning a CNN is a trade-off between efficiency and accuracy. We compare different methods by fixing one metric and comparing the other. 

On AlexNet, by achieving smaller accuracy loss (1.43\% ours vs. 2.00\% \cite{PerforatedCNN}), our method NISP-A manages to reduce significantly more FLOPs (67.85\%) than the one in \cite{PerforatedCNN} (50\%), denoted as ``Perforate" in the table; 
compare to the method in \cite{learning}  (denoted as ``Learning"), our method NISP-C achieves much smaller accuracy loss (0.54\% ours vs. 1.20\%) and prunes more FLOPs (53.70\% ours vs. 48.19\%). 
We manage to achieve 0 accuracy loss and reduce over 40\% FLOPs and 47.09\% parameters (NISP-D). 
On GoogLeNet, Our method achieves similar accuracy loss with larger FLOPs reduction (58.34\% vs. 51.50\%) 
% and higher GPU speedup (reduce GPU time by 30.12\% vs. 19.13\%) than the method in \cite{Tucker} (denoted as ``Tucker"). 
Using ResNet on Cifar10 dataset, with top-1 accuracy loss similar to \cite{pruneweigth} (56-A, 56-B. 110-A and 110-B), our method reduces more FLOPs and parameters. 

We also conduct our ResNet experiments on ImageNet \cite{imagenet_cvpr09}. We train a ResNet-34 and a ResNet-50 for 90 epochs. 
% The top-1 accuracy we obtain for the original ResNet-18 and ResNet-50 are 66.81\% and 72.91\% 
% following an open-source code\footnote{https://github.com/yihui-he/resnet-imagenet-caffe}
% (66.45\% was reported). 
For both ResNet models, we prune 15\% and 25\% of filters for each layer (denote as ``NISP-X-A" and ``NISP-X-B" (``X" indicates the ResNet model) in Table \ref{table:others}), and obtain 27-44\% FLOPs and parameter reduction with tiny top-1 accuracy loss, which shows superior performance when compared with the state-of-the-art methods \cite{pruneweigth,thinet}.

\subsection{Additional Analysis}
Below, we provide case studies and ablation analysis to help understand the proposed NISP pruning algorithm.

\textbf{Similar Predictive Power of Networks Before/After Pruning.} 
To check whether the pruned network performs similarly with the original network, we compare the final classification results of the original AlexNet and the pruned one with fine-tuning using the ILSVRC2012 validation set. 85.9\% of the top 1 predictions of the two networks agree with each other, and 95.1\% top 1 predictions of the pruned network can be found in the top 5 predictions of the original network.
The above experiments show that the network pruned by NISP performs similarly with the original one.

\begin{figure}
\centering     %%% not \center
\subfigure[LeNet Prune 75\% and 90\%]{\label{fig:LeNetQ}\includegraphics[height=3.5cm,width=40mm]{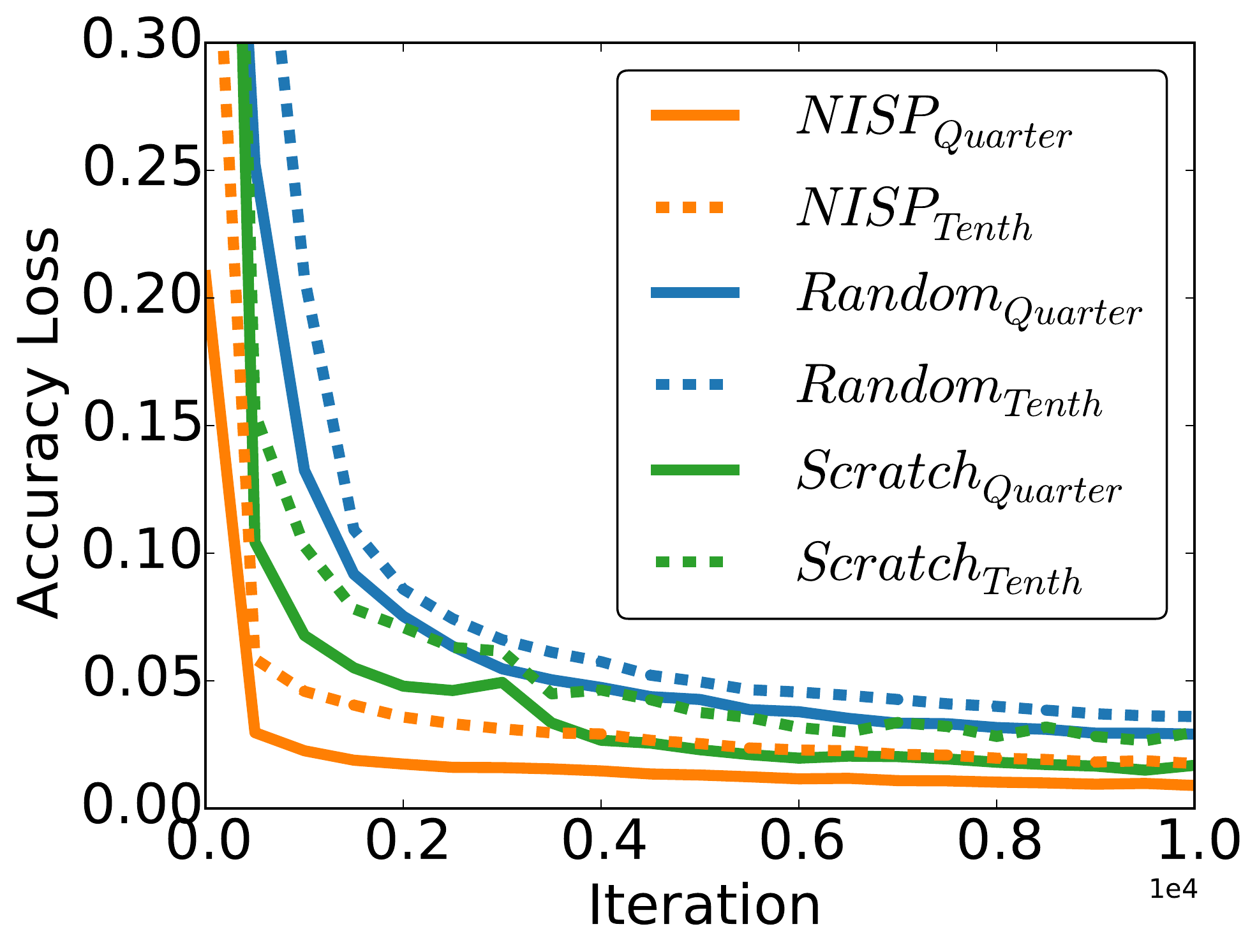}}
\subfigure[AlexNet Prune 75\%]{\label{fig:AlexQ}\includegraphics[height=3.5cm,width=40mm]{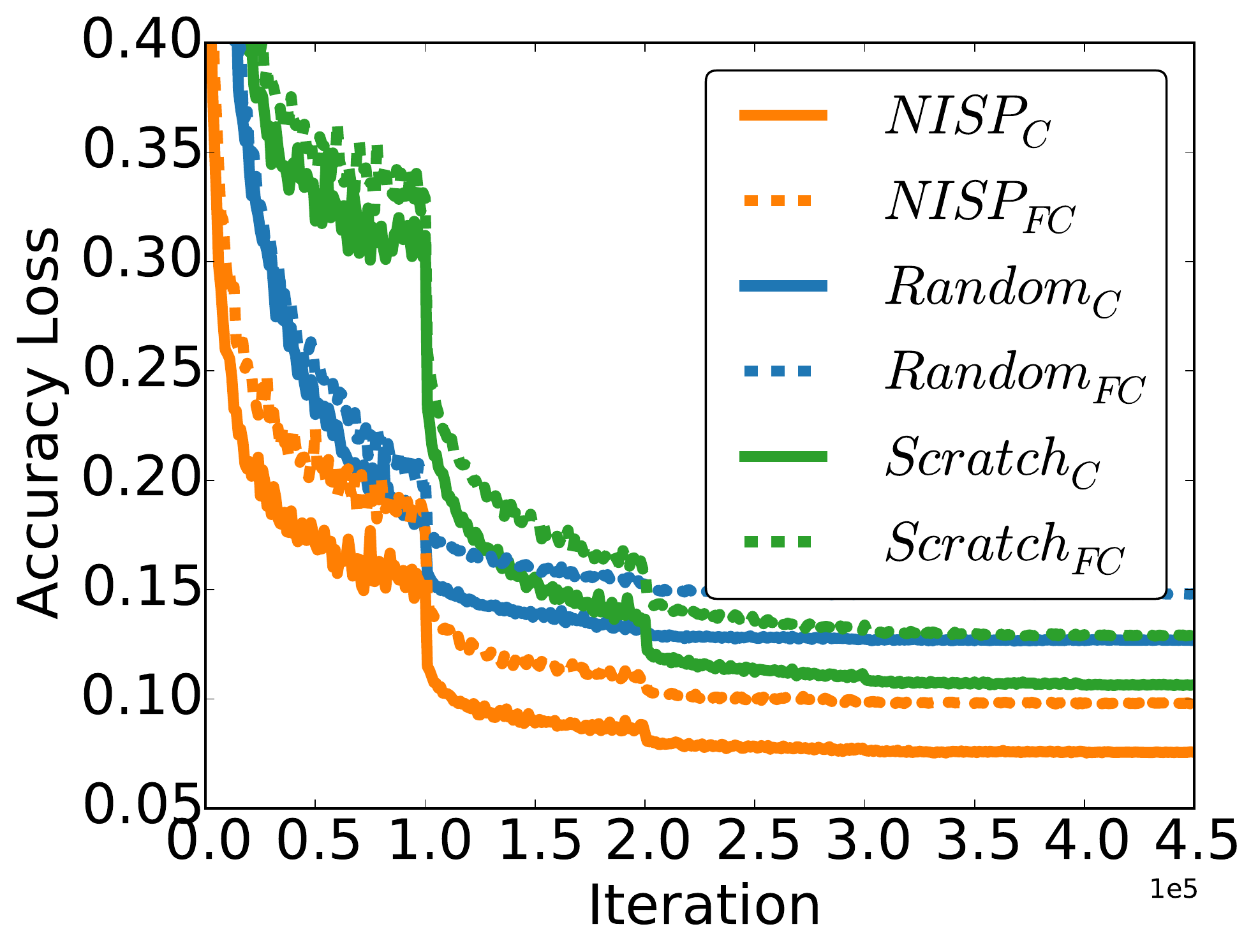}}

\caption{Evaluations for different pruning ratios (a) LeNet: pruning 75\% and 90\%, (b) AlexNet: pruning 75\%.  
CNNs pruned by NISP converge fastest with the lowest accuracy loss.}
\label{SuperALl}
\end{figure}

% \textbf{Recursive pruning.} One advantage of NISP is that it needs only a one-pass pruning with one-time fine-tuning, which is very efficient and it reduces significant computational cost. To study whether recursive pruning helps, we conduct a recursive pruning experiment on AlexNet. The final goal of the pruning ratio is 50\% for all layers. Instead of direct pruning, we prune 25\% of neurons prune twice using NISP. We observe that the importance score ranking of neurons generated by this recursive method slightly differs from the ranking list produced by the proposed direct approach, \eg, 
% %we compare the importance rankings of neurons in the final response layer between the one obtained before and after the first pruning, and we find that 
% among top 50\% most important neurons in both ranking lists, 84.27\% neurons overlap. The final accuracy change between direct pruning (76.38\% top-5 accuracy) and recursive pruning (76.71\%) is also small. We conclude that a neuron's importance score may change during recursive pruning, but after fine-tuning with a small learning rate, the change becomes negligible. 

\textbf{Sensitivity of pruning ratios.} 
% The main parameter of our method is the pruning ratio. We study the properties of different CNNs for different tasks, and provide experimental results to show the effectiveness of our method with different pruning ratios in the supplementary materials.
%The main parameter of our method is the pruning ratio. 
The selection of per-layer pruning ratios given a FLOPs budget is a challenging open problem with a large search space. Due to time limitation, we either choose a single pruning ratio for all layers or replicate the pruning ratios of baseline methods (\eg, \cite{Tucker}), and NISP achieves smaller accuracy loss, which shows the effectiveness of NISP. In practice, if time and GPU resources permit, one can search the optimal hyper-parameters by trying different pruning ratio combinations on a validation set.

% To study the effect of different pruning ratios, we experimented on pruning only a single layer or a type of layers by 50\% and monitored the accuracy loss. We observed that the sensitivities of pruning per layer are similar to each other for some networks (\eg, ResNet); some layers are more sensitive to a large pruning ratio (\eg, the Conv1 layer of AlexNet and the ``1x1" layers of GoogLeNet). 

We also evaluate NISP with very large pruning ratios. We test on pruning ratios of 75\% (denoted as \emph{Quarter} in the figures) and 90\% using LeNet (Fig. \ref{fig:LeNetQ}) (denoted as \emph{Tenth}) for both Conv and FC layers.
% The results are shown in Fig. .
For AlexNet (Fig. \ref{fig:AlexQ}), we test on pruning ratios of 75\% (\emph{Quarter}) for both convolution and FC layers, and we test two pruning strategies: (1) prune 75\% of neurons in FC layers and filters in Conv layers, denoted as \emph{$\text{FC}$}; and (2) only prune 75\% of the convolution filters without pruning FC layers, denoted as \emph{$\text{C}$}.

The above experiments show that NISP still outperforms all baselines significantly with large pruning ratios, in terms of both convergence speed and final accuracy.

\section{Conclusion}
We proposed a generic framework for network compression and acceleration based on identifying the importance levels of neurons. Neuron importance scores in the layer of interest (usually the last layer before classification) are obtained by feature ranking. We formulated the network pruning problem as a binary integer program and obtained a closed-form solution to a relaxed version of the formulation. We presented the Neuron Importance Score Propagation algorithm that efficiently propagates the importance to every neuron in the whole network. The network is pruned by removing less important neurons and fine-tuned to retain its predicative capability. Experiments demonstrated that our method effectively reduces CNN redundancy and achieves full-network acceleration and compression.

\section*{Acknowledgement}
The research was partially supported by the Office of Naval Research under Grant N000141612713: Visual Common Sense Reasoning for Multi-agent Activity Prediction and Recognition.

\section{Supplementary Material}

Despite their impressive predictive power on a wide range of tasks \cite{faster, xu1,xu3,yu1,yu2,peng,yu3,yu6,yu7,xu2,yu4,yu5}, the redundancy in the parameterization of deep learning models has been studied and demonstrated \cite{PredictingParameters}. We present NISP to efficiently propagate the importance scores from final responses to all other neurons to guide network pruning to achieve acceleration and compression of a deep network.
In the supplementary materials, we show details on how to propagate neuron importance from the final response layer, and some additional experiments.

\subsection{Neuron Importance Score Propagation (NISP)}
Given the importance of a neuron, we first identify the positions in the previous layer that are used as its input, then propagate the importance to the positions proportional to the weights. 
We only propagate the importance of the selected feature extractors to the previous layers and ignore the pruned ones.
The NISP process can be divided into three classes: from a 1-way tensor to a 1-way tensor, e.g. between FC layers; from a 1-way tensor to a 3-way tensor, e.g., from an FC layer to a conv/pooling layer; from a 3-way tensor to a 3-way tensor, e.g., from a pooling layer to a conv layer.

We simplify NISP by ignoring the propagation of bias.
% because there is no connection between cross-channel biases, and the importance.

\subsection{NISP: from 1-way tensor to 1-way tensor}

Given an FC layer with $M$ input neurons and $N$ output neurons, the $N
\text{-}by\text{-}1$ importance vector ($\mathbf{S}$) of the output feature is $\mathbf{S_{FC_{out}}}=\left [{S_{FC_{out}}}_1,{S_{FC_{out}}}_2 \dots {S_{FC_{out}}}_N\right ]^\text{T}$. 
We use $\mathbf{W_{FC}}\in \mathbb{R}^{M\times N}$ to denote the weights of the FC layer. The importance vector of the input neurons is:
\begin{equation}
\label{RI_FC_FC}
\mathbf{S_{FC_{in}}}=|\mathbf{W_{FC}}|  \cdot  \mathbf{S_{FC_{out}}}~,
\end{equation}
where $|\cdot|$ is element-wise absolute value.
\subsection{NISP: from 1-way tensor to 3-way tensor}
Given an FC layer with a 3-way tensor as input and $N$ output neurons, the input has a size of $X \times X \times C$, where $X$ is the spatial size and $C$ is the number of input channels. The input can be the response of a convolutional layer or a pooling layer.
We use $\mathbf{W_{FC}}\in \mathbb{R}^{(X^2 \times C)\times N}$ to denote the weights of the FC layer. The flattened importance vector $\mathbf{S_{in}} \in \mathbb{R}^ {(X^2 \times C)\times 1}$ of the input tensor is:
\begin{equation}
\label{RI_FC_conv}
\mathbf{S_{in}}=|\mathbf{W_{FC}}| \cdot \mathbf{S_{FC_{out}}}.
\end{equation}

\subsection{NISP: from 3-way tensor to 3-way tensor}
\subsubsection{Convolution Layer.}
% The RIP from a 3-way tensor to a 3-way tensor is much more complicated than the first two cases because of the large number of forward propagation operations between the input and output responses.
We derive NISP for a convolutional layer, which is the most complicated case of NISP between 3-way tensors.
NISP for pooling and local response normalization (LRN) can be derived similarly.

For a convolutional layer with the input 3-way tensor $\mathbf{{conv_{in}}} \in \mathbb{R}^ {X\times X \times N }$ and output tensor $\mathbf{{conv_{out}}} \in \mathbb{R}^ {Y\times Y \times F )}$, the filter size is $k$, stride is $s$ and the number of padded pixels is $p$.
During the forward propagation, convolution consists of multiple inner products between a kernel $\mathbf{k}_f \in \mathbb{R}^ {k\times k \times N}$, and multiple corresponding receptive cubes to produce an output response.
Fixing input channel $n$ and output channel $f$, the spatial convolutional kernel is $\mathbf{k}_{fn}$. For position $i$ in the $n^{th}$ channel of the input tensor, the corresponding response of the output channel $f$ at position $i$ is defined as Equation \ref{fw_conv}:
\begin{equation}
\label{fw_conv}
R_f(i)=\sum_n \mathbf{k}_{fn} \cdot \mathbf{in}(i),
\end{equation}
where $\mathbf{in}(i)$ is the corresponding 2-D receptive field. 
Given the importance cube of the output response $\mathbf{{S_{out}}} \in \mathbb{R}^ {Y \times Y \times F}$, we use a similar linear computation to propagate the importance from the output response to the input:
\begin{equation}
\label{bw_conv}
S_n(i)=\sum_f \mathbf{k}_{fn} \cdot \mathbf{S}_{out}(i),
\end{equation}
where $S_n(i)$ is the importance of position $i$ in the $n^{th}$ input channel, and $\mathbf{S}_{out}(i)$ is the corresponding 2-D matrix that contains the output positions whose responses come from the value of that input position during forward propagation.
We propagate the importance proportionally to the weights as described in Algorithm \ref{convBP}.
\begin{algorithm}[!t]
\caption{NISP: convolutional layer}\label{convBP}
\begin{algorithmic}[1]
\State $\mathbf{Input: }  \text{ weights of the conv layer } \mathbf{W} \in \mathbb{R}^ {X\times X \times N \times F} $
\State $\text{, flattened importance of the $f^{th}$ output channel }$ \State $\mathbf{S}_{out}^f \in \mathbb{R}^ {1 \times( X \times X )}$
\For {n in 1 \dots N}
\For {f in 1 \dots F}
\State $\mathbf{k}_{fn}  \gets |\mathbf{W}[:,:,n,f]|$
\State $\text{Construct } \mathbf{BP}_{conv}^{fn}$ as \eqref{BP_conv} and \eqref{b_c}
\State $\mathbf{S}_{in}^{fn} \gets \mathbf{S}_{out}^f \cdot
\mathbf{BP}_{conv}^{fn}$
\EndFor
\State $\mathbf{S}_{in}^{n} \gets \sum_f \mathbf{S}_{in}^{fn}$
\EndFor
\State $\mathbf{S}_{in} \gets [\mathbf{S}_{in}^{1},\mathbf{S}_{in}^{2} \dots, \mathbf{S}_{in}^{N}]$
\State \text{end}
\end{algorithmic}
\end{algorithm}

The propagation matrices used in algorithm \ref{convBP} are defined in \eqref{BP_conv} and \eqref{b_c}
\begin{equation}
\label{BP_conv}
\mathbf{BP}_{conv}^{fn}=\left[
\begin{aligned}
\mathbf{b}_1^{fn} \dots\ \  \mathbf{b}_j^{fn}  \ \ & \dots \,\mathbf{b}_k^{fn}  \\
\mathbf{b}_1^{fn} \dots\ \  &\mathbf{b}_j^{fn}  \ \  \dots \,\mathbf{b}_k^{fn}  \\
\vdots \\
\mathbf{b}_1^{fn} &\dots\ \  \mathbf{b}_j^{fn}  \ \ & \dots \,\mathbf{b}_k^{fn}  \\
\end{aligned}
\right],
\end{equation}
where $\mathbf{b}_c^i$ is the building block of size $Y \times X$ defined as:
\begin{equation}
\label{b_c}
\mathbf{b}_{i}^{fn}=\left[
\begin{aligned}
\mathbf{k}_{fn}[i,1] \dots\ \  & \dots \mathbf{k}_{fn}[i,k]   \\
\mathbf{k}_{fn}[i,1]&\dots \ \  \dots \mathbf{k}_{fn}[i,k]  \\
\vdots \\
&\mathbf{k}_{fn}[i,1]  \dots\ \   \dots \mathbf{k}_{fn}[i,k] 
\end{aligned}
\right],
\end{equation}

\begin{figure}[t]
\begin{center}
   \includegraphics[width=\linewidth]{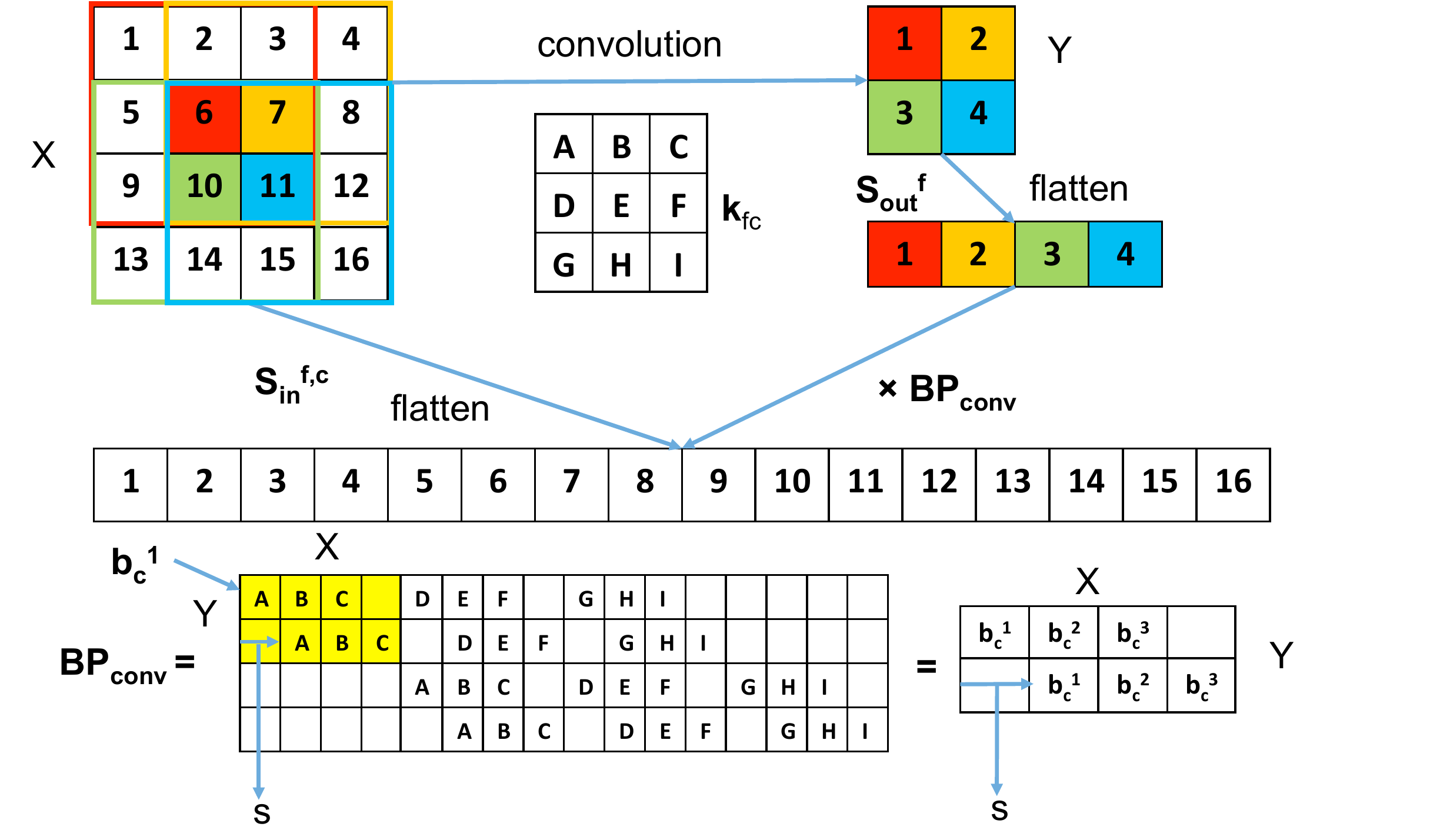}
\end{center}
   \caption{importance propagation: Convolutional layer. $X=4, Y=2, k=3, s=1$. Fixing the $f^{th}$ input channel and $c^{th}$ output channel, the upper-left X-by-X grid is the corresponding input feature map, and the upper-right Y-by-Y grid is the output map after convolution is applied. $\mathbf{k}_{fc}$ is the corresponding 2D convolutional kernel. Given the flattened importance vector for the output feature map $\mathbf{S_{out}}^{f,c}$, we use $\mathbf{BP}_{conv}$ to propagate the importance and obtain $\mathbf{S_{in}}^{f,c}$, which contains the importance of the input feature map. The structure of $\mathbf{BP}_{conv}$ is determined by the kernel size k and stride s. }
\label{fig:conv}
\end{figure}

Equation \ref{bw_conv} implies that the propagation of importance between 3-way tensors in convolutional layers can be decomposed into propagation between 2-D matrices. 
Fixing the input channel $n$ and the output channel $f$, the input layer size is $X \times X$ and the output size is $Y \times Y$. 
Given the flattened importance vector $\mathbf{S_{out}}^f \in \mathbb{R}^ {1\times (Y \times Y )}$ of the output layer, the propagation matrix $\mathbf{BP}_{conv}^{fn} \in \mathbb{R}^ { (Y \times Y ) \times (X \times X)}$ is used to map from $\mathbf{S_{out}}^f$ to the importance of input layer $\mathbf{S_{in}}^{fn} \in \mathbb{R}^ {1\times (X \times X )}$. 
$\mathbf{BP}^{fn}_{conv}(i,j)\neq0$,  implies that the $i^{th}$ position in the output layer comes from a convolution operation with the $j^{th}$ position in the input layer, and we propagate the importance between the two positions.
We use a $Y \times X$ matrix $\mathbf{b}^{fn}_i$ to represent the mapping between a row in the output layer to the corresponding row in the input layer.
In each row of $\mathbf{b}^{fn}_i$, there are $k$ non-zeros since each position in the output layer is obtained from a region with width $k$ of the input layer. The non-zeros of each row of $\mathbf{b}^{fn}_i$ are the $i^{th}$ row of the convolutional kernel $\mathbf{k}_{fn}$.
The offset of the beginning of the weights in each row is the stride $s$. 
The entire propagation matrix $\mathbf{BP}_{conv}^{fn}$ is a block matrix with each submatrix being a $Y \times X$ matrix of either $\mathbf{b}_i^{fn}$ or a zero matrix. 
Each row of $\mathbf{BP}_{conv}^{fn}$ has $\mathbf{b}_1^{fn}$ to $\mathbf{b}_k^{fn}$ because the height of a convolutional kernel is $k$. 
The offset of the beginning of the $\mathbf{b}$s in each row of $\mathbf{BP}_{conv}^{fn}$ is the stride s. 
We use the case when $X=4, Y=2, k=3, s=1$ as an example shown in Figure \ref{fig:conv}.

% \subsection{Other Layers}
% The RIP is difficult to compute between the output response and the input for the ReLU layer and Dropout layers because the operations are nonlinear. For a ReLU layer, since the ReLU operation is only conducted in isolation on a position without crossing channels or involving any other positions, we assume the importance can be propagated through the ReLU layer unchanged. Similarly, we propagate using the identity matrix for dropout layers because all neurons share the same dropout possibility.

% \subsection{importance propagation on Pooling and LRN Layers}
\subsubsection{Pooling Layer.}\label{pooling}
Assume a pooling layer with input tensor of size $X \times X \times F$ and output size $Y \times Y \times F$.
The pooling filter size is $k$ and the stride is $s$.
The basic idea of most pooling techniques is the same: use a fixed 2-dimensional filter to abstract local responses within each channel independently.
For example, in max pooling each output response consists of the max of $k \times k$ values from the input responses.
Due to the large variance of input data, it is safe to assume a uniform distribution on which value within the receptive field is the largest is a uniform distribution.
Consequently, for an output response location, the contributions from the corresponding $k \times k$ values of the input response are equal. 
Since pooling is a spatial operation that does not cross channels, we can propagate the importance of each channel independently. 
Given a flattened importance vector of a channel $f$ $\mathbf{{S_{out}}^f} \in \mathbb{R}^ {1\times (Y \times Y )}$ of the output 3-way tensor, the flattened importance vector of the input tensor is calculated as:

\begin{equation}
\label{RI_Pooling_Conv}
\mathbf{{S_{in}}^f}=\mathbf{{S_{out}}^f}\cdot \mathbf{BP}_{pooling},
\end{equation}
where $\mathbf{BP}_{pooling}$ is the back-propagation matrix of size $Y^2\times X^2$ defined as:
\begin{equation}
\label{BP_pooling}
\mathbf{BP}_{pooling}=\left[
\begin{aligned}
\mathbf{b}_p \dots\ \ \mathbf{b}_p \ \ &  \dots \,\mathbf{b}_p  \\
\mathbf{b}_p \dots\ \ &\mathbf{b}_p \ \  \dots \,\mathbf{b}_p \\
\vdots \\
\mathbf{b}_p & \dots\ \ \mathbf{b}_p \ \  \dots \,\mathbf{b}_p
\end{aligned}
\right],
\end{equation}
where $\mathbf{b}_p$ is the building block of size $Y \times X$ defined as:
\begin{equation}
\label{b_p}
\mathbf{b}_{p}=\left[
\begin{aligned}
1 \dots\ \ 1 \ \ & \dots \,1  \\
1 \dots\ \ &1 \ \  \dots \,1 \\
\vdots \\
1 & \dots\ \ 1 \ \  \dots \,1
\end{aligned}
\right],
\end{equation}

\begin{figure}[t]
\begin{center}
   \includegraphics[width=\linewidth]{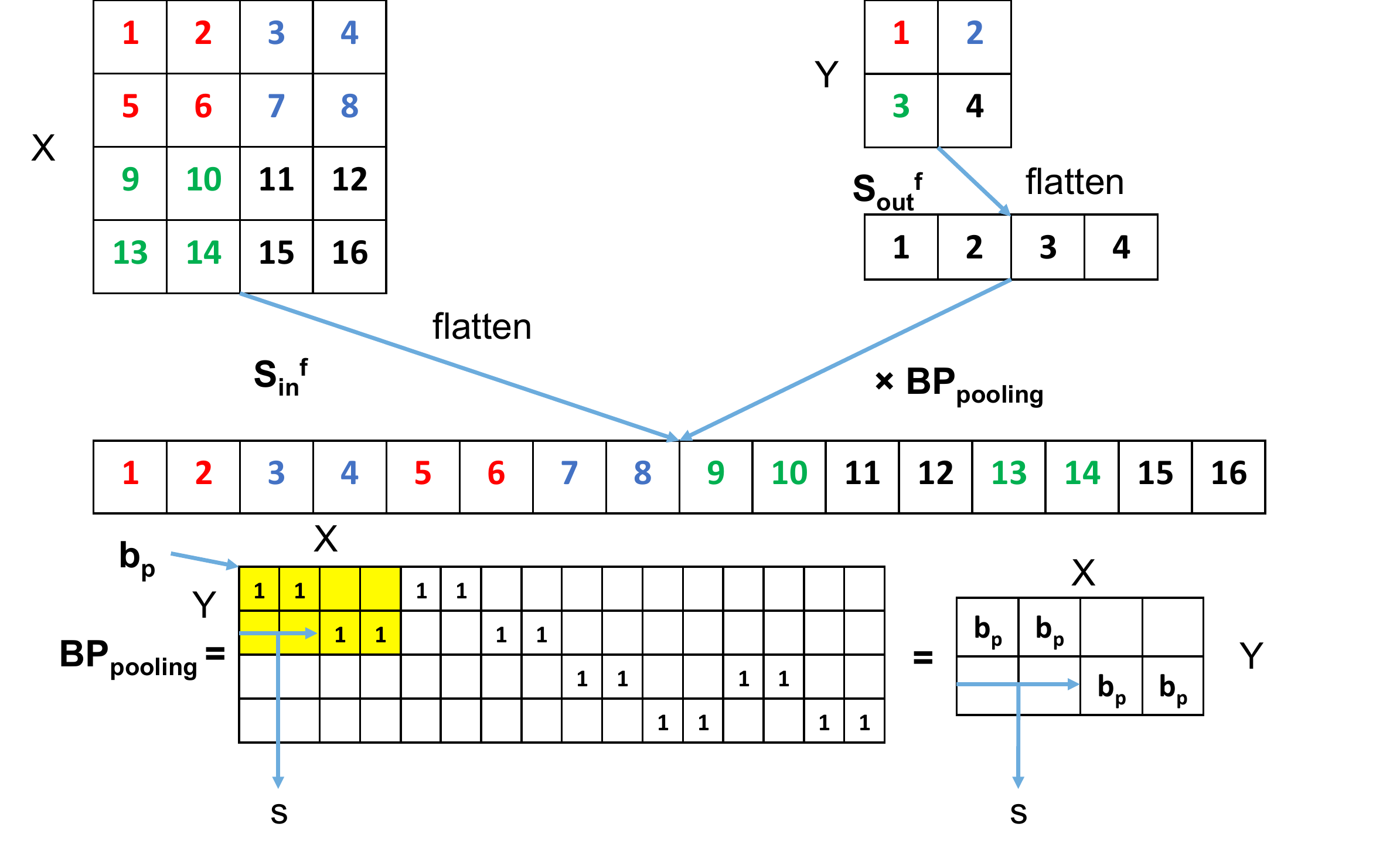}
\end{center}
   \caption{NISP: Pooling layer. $X=4, Y=2, k=2, s=2$. The upper-left X-by-X grid is the $f^{th}$ feature map of the input channel, and the upper-right Y-by-Y grid is the output channel after pooling is applied. Given the importance vector $\mathbf{S_{out}}^f$, we use $\mathbf{BP}_{pooling}$ to propagate the importance and obtain $\mathbf{S_{in}}^f$, which contains the importance of each position of the input feature map. The structure of $\mathbf{BP}_{pooling}$ relates to the kernel size k and stride s.}
\label{fig:long}
\end{figure}

Consider one channel with input size $X \times X$ and the output size $Y \times Y$.
Given the flattened importance vector $\mathbf{S_{out}}^f \in \mathbb{R}^ {1\times (Y \times Y )}$ of the output layer, the propagation matrix $\mathbf{BP}_{pooling} \in \mathbb{R}^ { (Y \times Y ) \times (X \times X)}$ is used to map from $\mathbf{S_{out}}^f$ to the importance of input layer $\mathbf{S_{in}}^f \in \mathbb{R}^ {1\times (X \times X )}$.
If $\mathbf{BP}_{pooling}(i,j)=1$, the $i^{th}$ position in the output layer comes from a pooling operation and involves the $j^{th}$ position in the input layer, so we propagate the importance between the two positions.
We use a $Y \times X$ matrix $\mathbf{b}_p$ to represent the mapping between a row in the output layer to the corresponding row in the input layer.
In each row of $\mathbf{b}_p$, there are $k$ $1's$ since each element in the output layer is pooled from a region with width $k$ of the input layer. 
The offset of the beginning of the $1's$ is the stride $s$. 
The entire propagation matrix $\mathbf{BP}_{pooling}$ is a block matrix with each submatrix being a $Y \times X$ matrix of either $\mathbf{b}_p$ or a zero matrix. 
Each row of $\mathbf{BP}_{pooling}$ has $k$ $\mathbf{b}_p s$ because the height of pooling filter is $k$. 
The offset of the beginning of the $k$ $\mathbf{b}_p$s is the stride s.
The ones in $\mathbf{b}_p$ will be normalized by the number of positions covered by a pooling filter (the same for LRN layers shown below). 
The other elements are all zeros. 
We use the case that $X=4, Y=2, k=2, s=2$ as an example shown in Figure \ref{fig:long}.

\subsubsection{Local Response Normalization Layer.}
Krizhevsky \emph{et al.} \cite{Alexnet} proposed Local Response Normalization (LRN) to improve CNN generalization.
For cross-channel LRN, sums over adjacent kernel maps at the same spatial position produce a response-normalized activation at that position. 
Since LRN is a non-linear operation, it is intractable to conduct exact importance propagation between the input and output tensors. 
One way to approximate propagation is to assume the kernel maps at one spatial position contribute equally to the response at that position of the output tensor when considering the large variance of the input data. 
Then, given the $X\times X \times N$ importance tensor for the response of a LRN layer with $\text{local}\_\text{size} = l$, which is the number of adjacent kernel maps summed for a spatial position, considering all $N$ channels of a spatial position $(i,j)$, the importance vector of that spatial position is $\mathbf{S}_{out}^{ij} \in \mathbb{R}^{1 \times N}$. The corresponding importance vector of the input $\mathbf{S}_{in}^{ij} \in \mathbb{R}^{1 \times N}$ is: 
\begin{equation}
\label{lrn}
\mathbf{S}_{in}^{ij}=\mathbf{S}_{out}^{ij} \cdot \mathbf{BP}_{LRN},
\end{equation}
where $\mathbf{BP}_{LRN} \in \mathbb{R}^{N \times N}$ is defined as:
% \begin{spacing}{0.8}
\begin{equation}
\label{BP_LRN}
\mathbf{BP}_{LRN}=\left[
\renewcommand{\arraystretch}{0.6}
\setlength{\arraycolsep}{1pt}
\begin{array}{cccccccccccc}
&1  &1  &\cdots  &1 & & & & & & &  \\
&1  &1  &\cdots &1 &1  & & & & & &  \\
&\cdot &\cdot  &\cdot  &\cdot  &\cdot  &\cdot & & & & &  \\
& 1 &1  &\cdots &1 &1 &\cdots  & & & & & \\
& & 1 &\cdots &1 &1 &\cdots &1 & & & &\\
& & & &\cdot &\cdot &\cdot &1 &1 & & &\\
& & & &1 &1 &\cdots &\cdots &\cdots & & &\\
& & & & &1 &\cdots &1 &1 &\cdots & 1 & \\
& & & & & & &1  &1  &\cdots &1 &1 \\
& & & & & & &\cdot  &\cdot  &\cdot  &\cdot  &\cdot  \\
& & & & & & &1  &1  &\cdots &1 &1 \\
& & & & & & &  &1  &\cdots &1 &1 \\
\end{array}
\right].
\end{equation}
% \end{spacing}

\begin{figure}[t]
\begin{center}
   \includegraphics[width=0.95\linewidth]{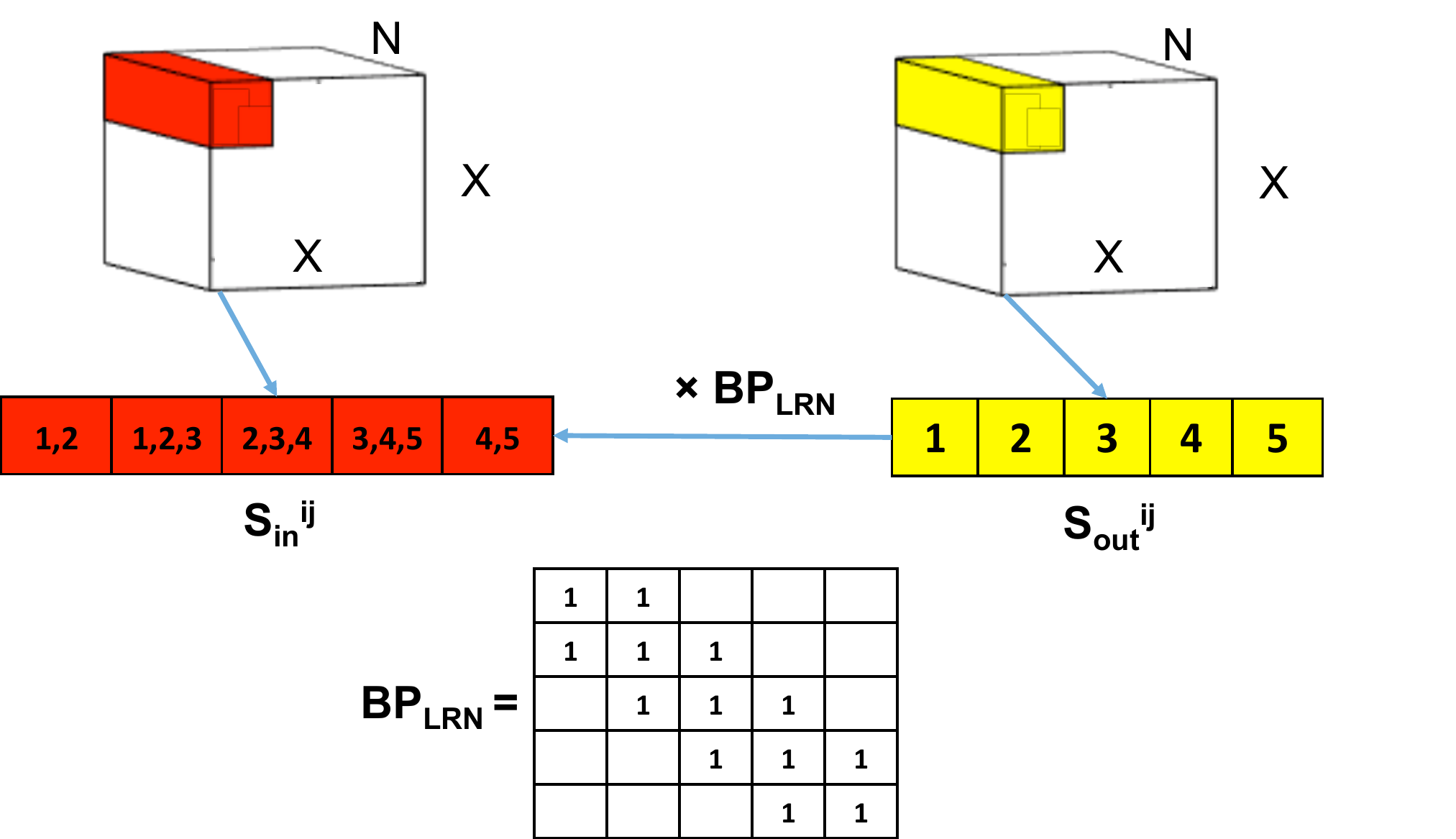}
\end{center}
   \caption{Importance propagation: LRN layer (cross-channel). $l=3, N=5$. The red vector is the cross-channel vector at spatial position $(i,j)$ of the input tensor, and the yellow vector is the cross-channel vector at the same position of the output tensor after LRN is applied. Given the $\mathbf{S_{out}}^{ij}$, we use $\mathbf{BP}_{LRN}$ to propagate the importance and obtain $\mathbf{S_{in}}^{ij}$, which contains the importance of each position of the input feature map. The structure of $\mathbf{BP}_{LRN}$ relates to the local size l and number of channels N.}
\label{fig:lrn}
\end{figure}

For a cross-channel LRN, the output response tensor has the same shape as the input. 
For a spatial position $(i,j)$ of the output tensor, given its importance vector $\mathbf{S}_{out}^{ij}$, we construct a $N \times N$ symmetric matrix $\mathbf{BP}_{LRN}$ to propagate its importance to the corresponding input vector$\mathbf{S}_{in}^{ij}$ at that position.
Since the center of the LRN operation is at position $(i,j)$, the operation will cover $\frac{l+1}{2}$ positions to the left and right.
When the operation is conducted on the positions at the center of the vector $\mathbf{S}_{out}^{ij}$ (from column $\frac{l+1}{2}$ to $N\text{-}\frac{l+1}{2}+1$), the operation covers $l$ cross-channel position so that the corresponding columns in $\mathbf{BP}_{LRN}$ have $l$ 1's. 
When the LRN operation is conducted at the margin of the vector, there are missing cross-channel positions so that from column $\frac{l+1}{2}$ to column 1 (similar for the right-bottom corner), the 1's in the corresponding column of $\mathbf{BP}_{LRN}$ decreases by 1 per step from the center to the margin. 
We use the case when $l=3, N=5$ as an example of LRN layer with cross-channel in Figure \ref{fig:lrn}. 

For within-channel LRN, following our equal distribution assumption, the importance can be propagated similarly as in a pooling layer.

\subsection{Experiments}

\subsection{PCA Accumulated Energy Analysis}
\begin{figure}[!t]
\centering     %%% not \center
\subfigure[LeNet]{\label{fig:PCAL}\includegraphics[width=.49\linewidth]{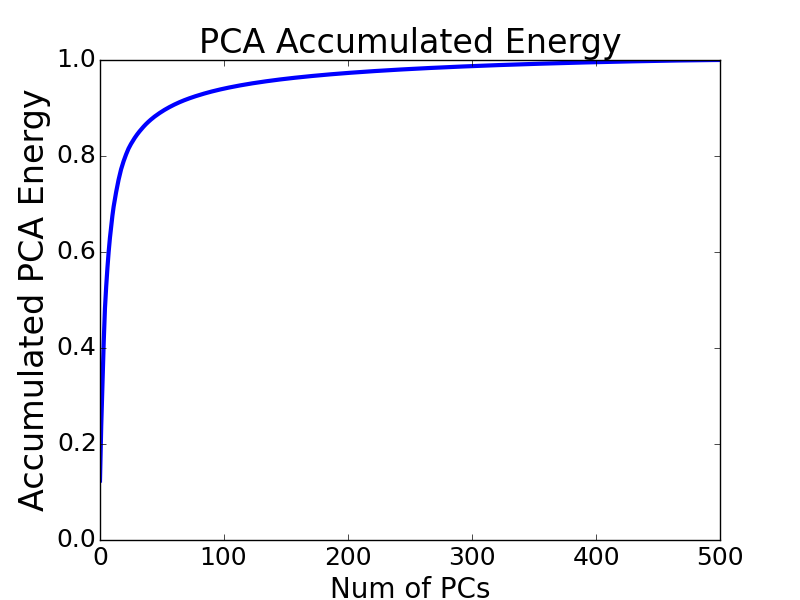}}
\subfigure[AlexNet]{\label{fig:PCAA}\includegraphics[width=.49\linewidth]{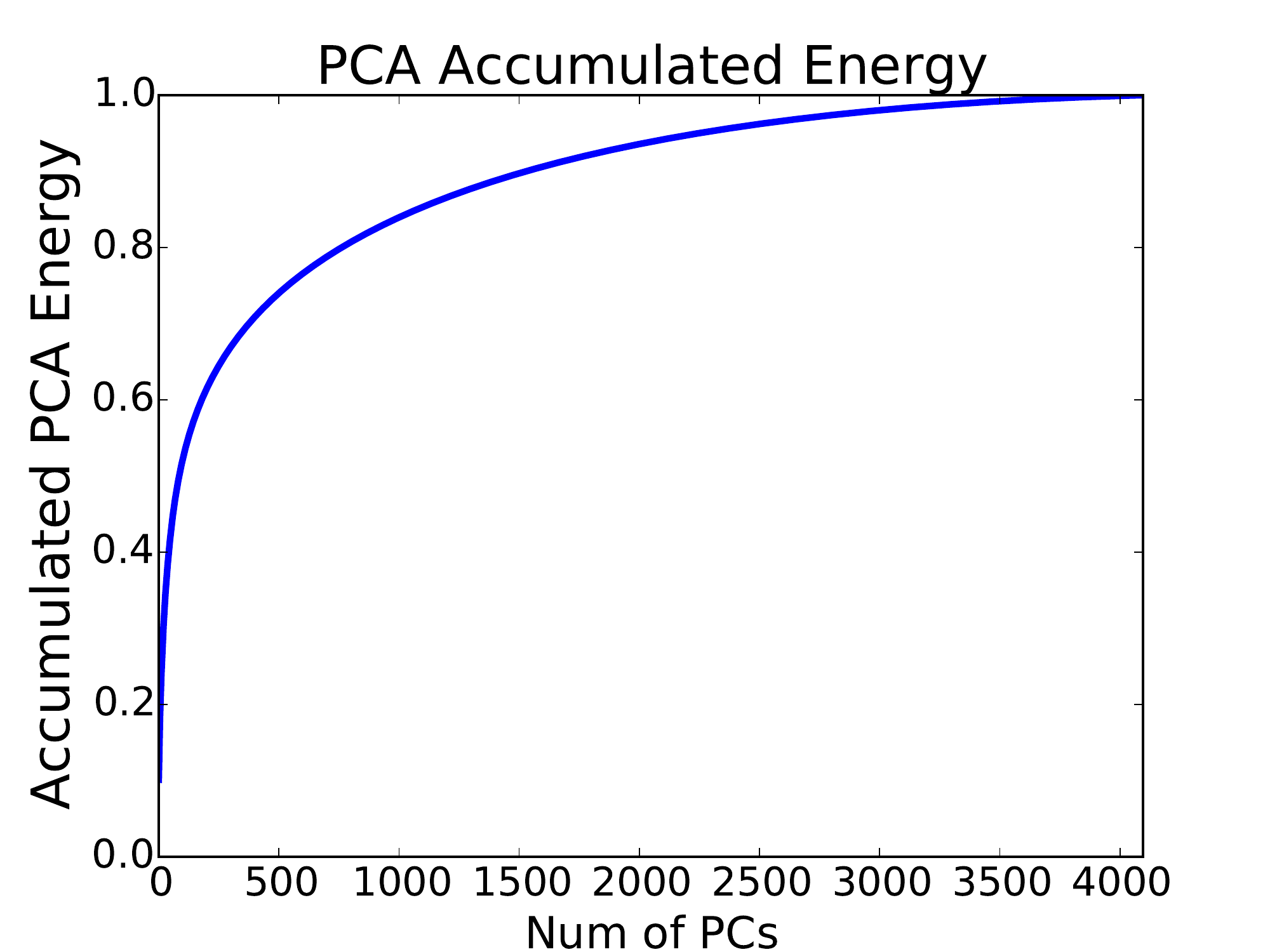}}
\caption{PCA accumulated energy analysis: LeNet on MNIST (a) and AlexNet on ImageNet (b). The y axis measures the PCA accumulated energy. The x axis shows the number of PCs.}
\end{figure}

One way to guide the selection of pruning ratio is the PCA accumulated energy analysis \cite{Nonlinear} on the responses of a pre-pruned layer. The PCA accumulated energy analysis shows how many PCs it needs for that layer to capture the majority of variance of the samples, which implies a proper range of how many neurons/kernels we should keep for that layer. We show the PCA accumulated energy analysis results on the last FC layers before the classification part for LeNet (ip1) and AlexNet (fc7) in Figure \ref{fig:PCAL} and \ref{fig:PCAA}. By setting variance threshold as 0.95, 120 out of 500 PCs are required for LeNet, 2234 out of 4096 PCS are required for AlexNet to capture the variance.

\subsection{Experiments on AlexNet: Convolutional Layers v.s. FC Layers}
\begin{figure}[!t]
\begin{center}
   \includegraphics[width=.9\linewidth]{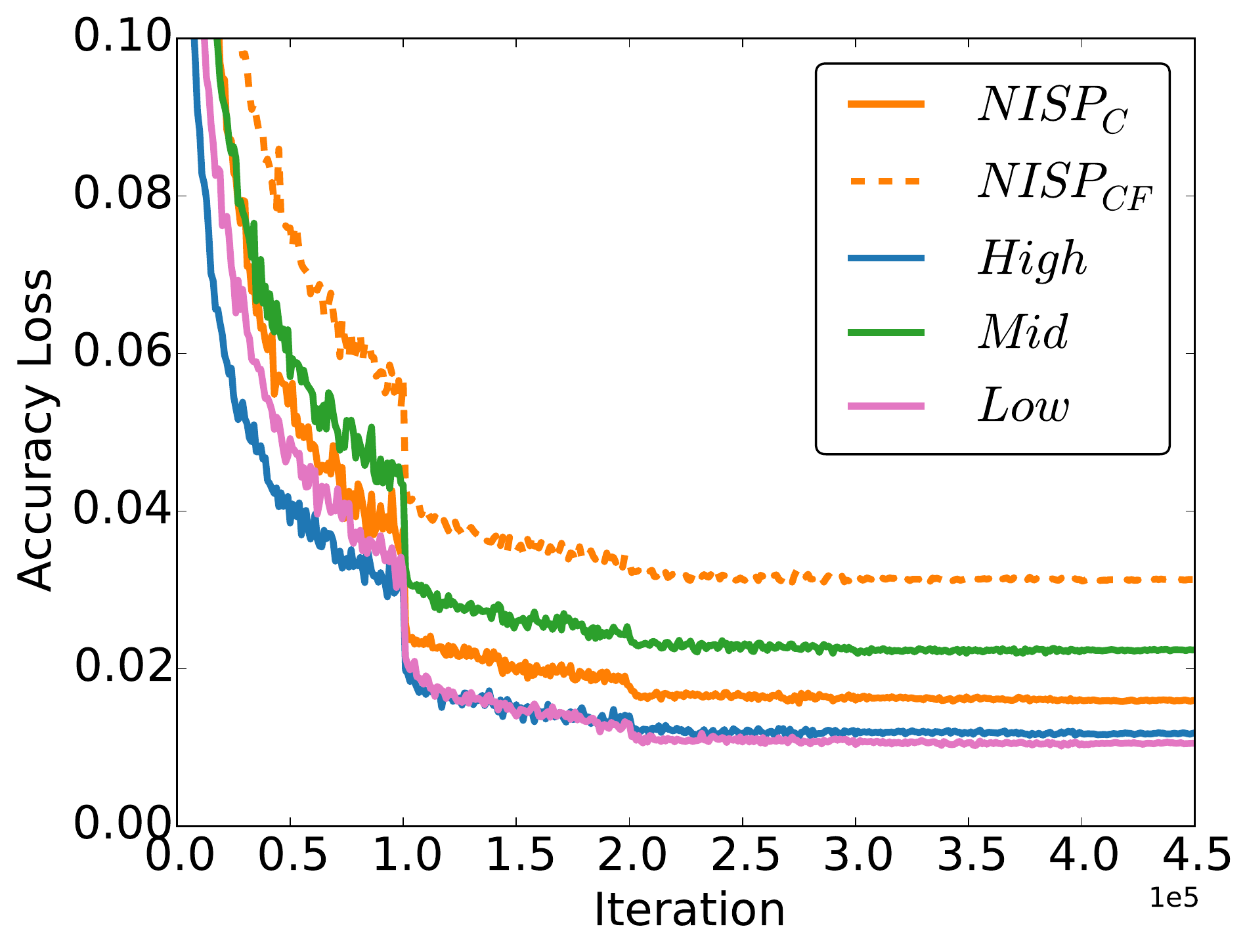}
\end{center}
   \caption{Learning Curves of AlexNet on ImageNet: The subscript $CF$ means we prune both convolutional kernels and neurons in FC layers, and $C$ means we only prune convolutional kernels. \emph{$\text{High}$}, \emph{$\text{Mid}$} and \emph{$\text{Low}$} mean we prune the entire CNN except for the high/middle/low level convolutional layers (Conv4-Conv5, Conv3 and Conv1-Conv2 respectively). }
\label{fig:alexThree}
\end{figure}

%As shown in section \ref{baseline}, our proposed IS based method significantly outperforms the baselines. 
From the experiments in the main paper, we found that FC layers have significant influence on accuracy loss, model size and memory usage. 
To exploit the impact of pruning FC layers and convolutional layers, we conduct experiments on pruning half of the neurons in FC layers and some convolutional layers. 
We categorize the 5 convolutional layers into three-level feature extractors: low (Conv1-Conv2 layers), middle (Conv3 layer) and high (Conv4-Conv5 layers).
Figure \ref{fig:alexThree} displays learning curves and shows that although FC layers are important in AlexNet, powerful local feature extractors (more kernels in convolutional layers) can compensate the loss from pruning neurons in FC layers, or even achieve better predictive power (High and Low curves). 
% The benchmark of the CNNs pruned based on the above strategies are shown from column 2 to 6 of Table~\ref{tableAlex}.

\subsection{Experiments on GoogLeNet}
\begin{figure}[!t]
\begin{center}
  \includegraphics[width=.9\linewidth]{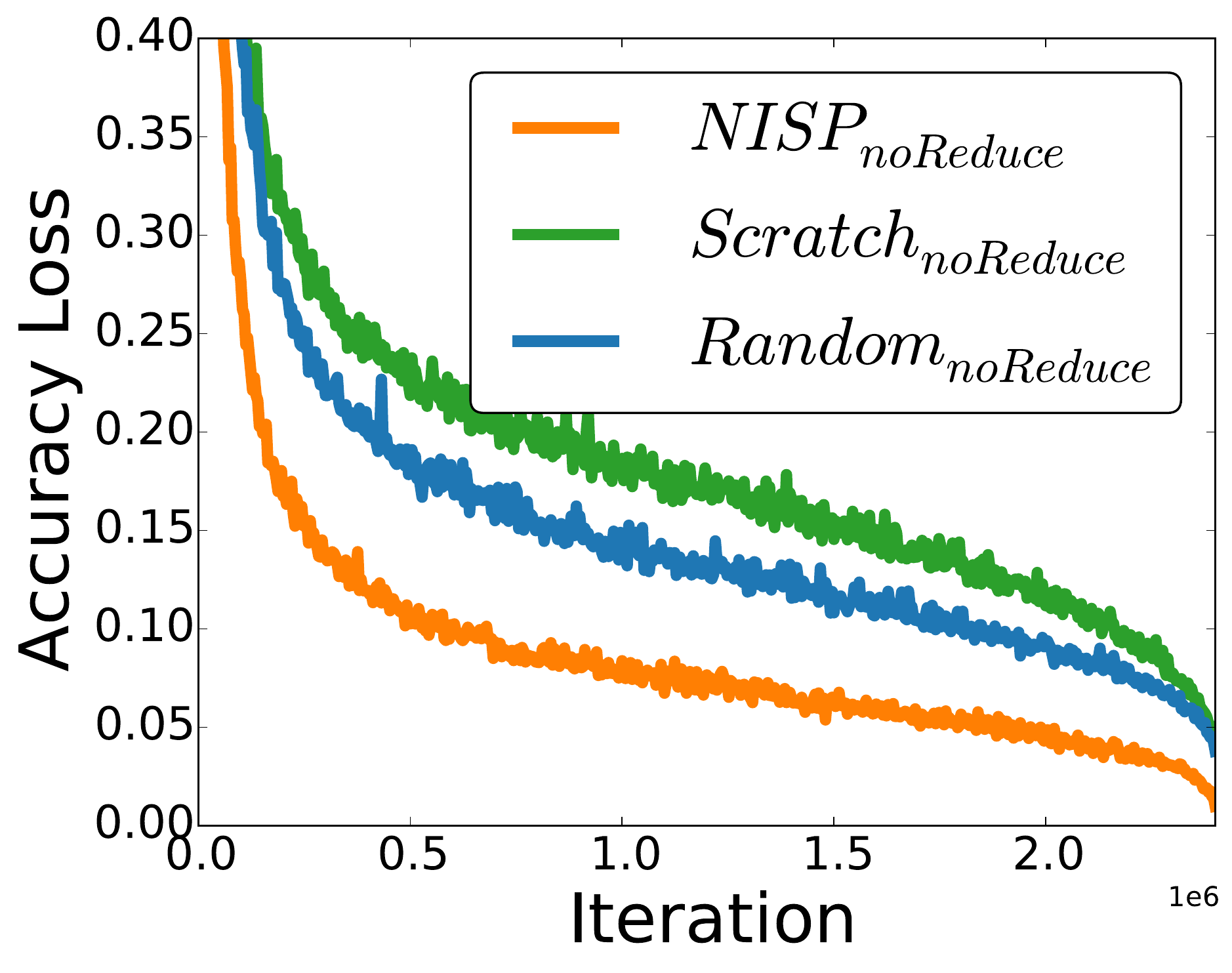}
\end{center}
  \caption{Learning Curves of GoogLeNet on ImageNet: The pruning ratio is 50\%. We prune all layers but the reduction layers in the inception modules. importance based pruning method converges much faster and can achieve the smallest accuracy loss. }
\label{fig:googRe}
\end{figure}

The learning curves for ``no\_Reduce" is shown in Figure \ref{fig:googRe}. We observe that our importance based pruning method leads to better initialization, faster convergence and smaller final accuracy loss.

\subsection{Layer-wise Improvements}
\begin{figure}[!t]
\begin{center}
  \includegraphics[width=\linewidth]{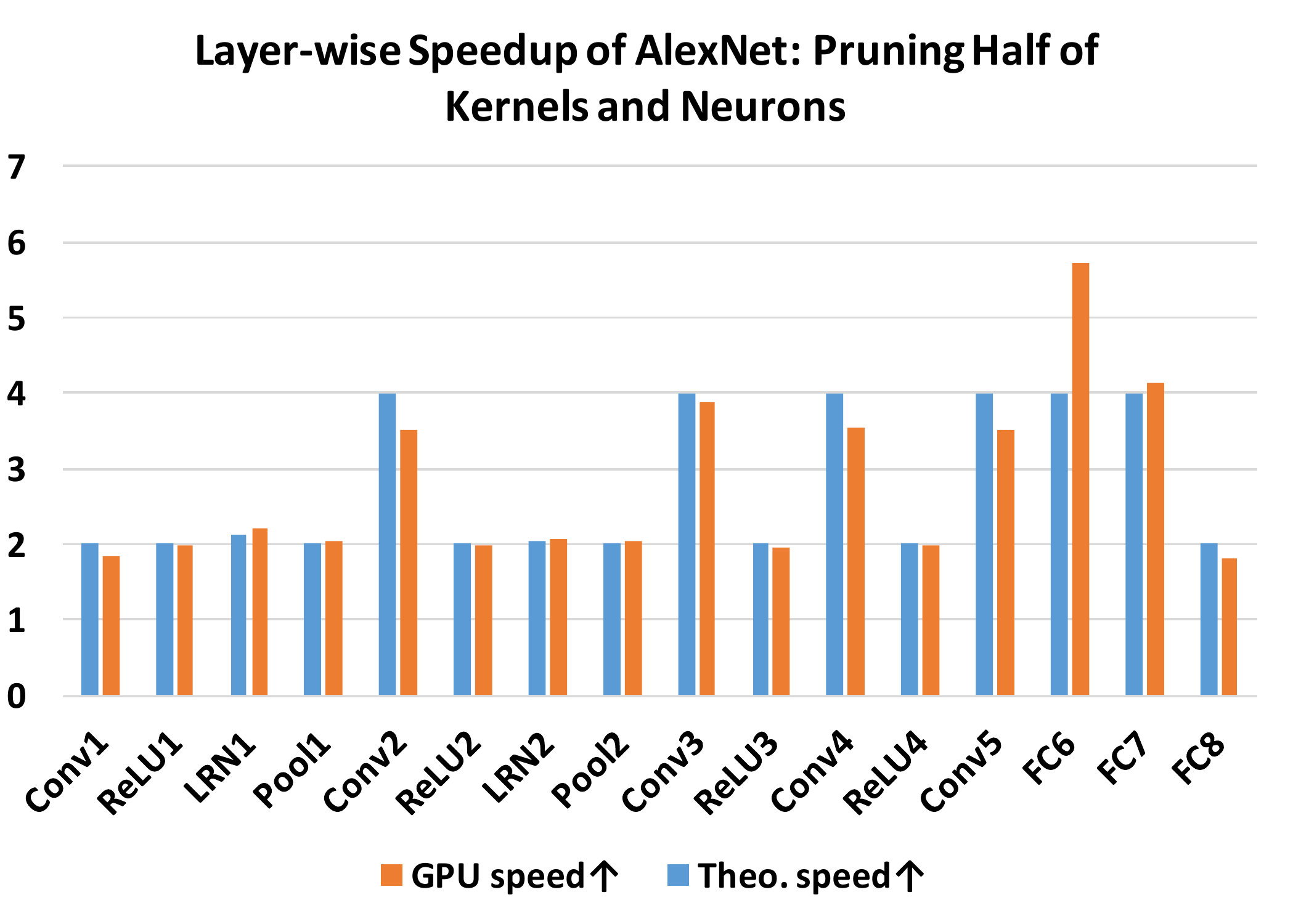}
\end{center}
  \caption{Full-Network Acceleration of AlexNet: Pruning Half of the Kernels and Neurons. }
\label{fig:layerwise}
\end{figure}

In our experiments of AlexNet on Titan X, the empirical computation time for the intermediate layers (all layers except for convolutional layers and FC layers) accounts for 17\% of the entire testing time; therefore, those layers must be considered as well while designing an acceleration method.
One of our advantages over existing methods is that all layers in the network can be sped up due to the fact that the data volume or feature dimension at every layer is reduced.
For example, by pruning kernels in convolutional layers, we reduce the number of both output channels of the current layer and input channels of the next layer. 
In theory, given a pruning ratio of 50\%, except for the first layer whose input channels cannot be pruned, all of the convolutional layers can be sped up by $4\times$. 
The intermediate pooling, non-linearity and normalization layers have a theoretical speedup ratio of around $2\times$. 
The layer-wise acceleration ratios (both theoretical and empirical) of our method when the pruning ratio is 50\% for both convolutional layers and FC layers are shown in Figure \ref{fig:layerwise}. 
We observe that the theoretical and empirical speedup are almost the same for pooling, non-linearity and normalization.
% The acceleration of entire networks helps the empirical speedup ratio ($2.80\times$) approach the theoretical speedup ratio ($3.46\times$).

{
\bibliography{egbib}
\bibliographystyle{ieee}
}

\end{document}